## IMPORTANT INFORMATION

This dissertation was written and submitted by **Mandar Marathe** in partial fulfilment of the requirements for the Master of Arts degree in Advanced Arabic at the **University of Exeter** in September 2022.

The dissertation represents an early conceptual analysis and preliminary feasibility study for the quantitative assessment of Arabic rhetoric, based on the relative frequency analysis of rhetorical devices within a text. Conceptually, this is a method which is equally applicable to Arabic rhetoric as it is to rhetoric in other languages.

The ideas and methods presented here have since been significantly improved and expanded in subsequent work, resulting in the development of two novel quantitative linguistics metrics: **Rhetorical Density** (or *Rhetoric Density*) and **Rhetorical Diversity**, which are the foundation of the **BALAGHA Score**.

This dissertation has been made publicly available (DOI: **10.5281/zenodo.15765533**) as it is the first formal description and documentation of early concepts that would subsequently become Rhetorical Density, Rhetorical Diversity, and the BALAGHA Score.

Please visit **https://marathe.org/ma-dissertation** for the latest updates, publications, and ongoing research in relation to Rhetorical Density, Rhetorical Diversity, and the BALAGHA Score.

Mandar Marathe
mandar@marathe.org
https://marathe.org
https://orcid.org/0000-0002-6927-6836
29 June 2025

# Creation of a Numerical Scoring System to Objectively Measure and Compare the Level of Rhetoric in Arabic Texts: A Feasibility Study, and A Working Prototype

**MA Advanced Arabic**

**ARAM027 Dissertation**

**Mandar Marathe**

**mm1126@exeter.ac.uk**

**September 2022**

**Supervisor: Dr Mustafa Baig**

Word count: 16,302

## Abstract

Arabic Rhetoric is the field of Arabic linguistics which governs the art and science of conveying a message with greater beauty, impact and persuasiveness. The field is as ancient as the Arabic language itself and is found extensively in classical and contemporary Arabic poetry, free verse and prose. In practical terms, it is the intelligent use of word order, figurative speech and linguistic embellishments to enhance message delivery. Despite the volumes that have been written about it and the high status accorded to it, there is no way to objectively know whether a speaker or writer has used Arabic rhetoric in a given text, to what extent, and why. There is no objective way to compare the use of Arabic rhetoric across genres, authors or epochs. It is impossible to know which of pre-Islamic poetry, Andalucian Arabic poetry, or modern literary genres are richer in Arabic rhetoric. The aim of the current study was to devise a way to measure the density of the literary devices which constitute Arabic rhetoric in a given text, as a proxy marker for Arabic rhetoric itself. A comprehensive list of 84 of the commonest literary devices and their definitions was compiled. A system of identifying literary devices in texts was constructed. A method of calculating the density of literary devices based on the morpheme count of the text was utilised. Four electronic tools and an analogue tool were created to support the calculation of an Arabic text's rhetorical literary device density, including a website and online calculator. Additionally, a technique of reporting the distribution of literary devices used across the three sub-domains of Arabic rhetoric was created. The output of this project is a working tool which can accurately report the density of Arabic rhetoric in any Arabic text or speech.

# Table of Contents

## List of Figures

## List of Tables

## Translation, Transliteration and Referencing

Translation and transliteration follow the International Journal of Middle East Studies Translation and Transliteration Guide.[1] Referencing follows the Chicago Manual of Style.

[1] "IJMES Translation and Transliteration Guide," IJMES, accessed August 29, 2022, https://www.cambridge.org/core/services/aop-file-manager/file/630d1d754cf5970011527e65/IJMES-Transliteration-Guide.pdf.

# Abbreviations

| | |
|---|---|
| AR | Arabic rhetoric (*al-balāgha*) |
| ARLDDMS | Arabic Rhetoric Literary Device Density Measurement System |
| ARLDDS | Arabic Rhetoric Literary Device Density Score |
| ARLDDS | Arabic Rhetoric Literary Device Density Summary |
| AR LD | Arabic rhetoric literary devices |
| LD | Literary device |

# Definitions

| | |
|---|---|
| Addressee | Any entity that is receiving the speech or text produced by the communicator. In Arabic-language AR texts, often referred to as *al-mukhāṭib* or *al-sāmiʿ* or *al-qārīʾ*. |
| Assessor | Any person who is assessing a sample of Arabic text or speech with the objective of realising the aims of the ARLDDMS. |
| Communicator | Any entity that produces oral or written speech, such as the speaker or the writer. In Arabic-language AR texts, often referred to as *al-mutakallim* or *al-khaṭīb*. |
| Lexical item | A full or partial utterance, collection of words, sentence or phrase which is usually smaller than the entire proposition. |
| Literary device | A linguistic feature such as metaphor, simile, allegory or alliteration. |
| Morpheme | The smallest indivisible lexical item in the language. In addition to nouns and verbs, it includes prefixes, suffixes and pronouns and conjunctions. |
| Proposition | A full thought or piece of information conveyed from the communicator to the addressee. |
| Text | Any spoken or written communication produced by the communicator. The word "text" is used for convenience and includes oral output as well as written text. |

# Chapter 1
# Introduction

## 1.1) Introduction to Arabic rhetoric

In the field of Arabic linguistics, the term *balāgha* refers to the science associated with the "study of aesthetic effectiveness," or the "conveying of meaning in the best of verbal forms."[2] It is often translated as "rhetoric" because it may be used to enhance persuasiveness of speech, although it does not have the negative connotations associated with the English word "rhetoric." Arabic rhetoric (*balāgha*, AR) overlaps with "eloquence" (*faṣaha*) which refers to the clarity of speech, and is a distinct field in Arabic linguistics alongside grammar, morphology, stylistics, logic and poetics. Abdul-Raof describes AR as the "flesh and blood of the Arabic language."[3] In simple terms, it is the use of literary devices (LDs) such as metaphor, simile, alliteration, and modifications to word order to change the meaning of the sentence, as a means to more effective, engaging and efficient communication.

AR has existed since before the era of oral pre-Islamic poetry (prior to approximately 650 AD) up to the present day. The oral sciences of this field underwent a process of classification, consolidation and written documentation between 750 AD and 1400 AD which resulted in a broad classification of AR into three sub-domains that remains to the present day: (i) the science of word order and sentence construction (*'ilm al-ma'ānī*), (ii) the science of figurative speech (*'ilm al-bayān*), (iii) and the science of linguistic embellishments (*'ilm al-badī'*).[4]

Arabic literature has undergone tremendous change since the era of pre-Islamic poetry to the modern age. The Arab Awakening (*an-nahḍa*) of the late 19th century to the early 20th century was a pivotal stage which saw an accelerated rate of change resulting in contemporary Western genres of literature taking precedence over the classical forms of Arabic poetry. Modern Arabic literature now mirrors and competes with all the genres and styles present in other languages, cultures and societies.

---

[2] Kate Zebiri, "Towards a Rhetorical Criticism of the Qur'an," *Journal of Qur'anic Studies* 5, 2 (2003): 95-120.

[3] Hussein Abdul-Raof, *Arabic Rhetoric: A Pragmatic Analysis*, (Oxford: Routledge, 2006), i.

[4] H. Abdul-Raof, 2006, 31-75.

While AR has been presented as a vital linguistic and communicative skill, the question which arises is how has AR fared during the modernisation of Arabic literature? We have seen the venerable and seemingly timeless *qaṣīda* form of Arabic poetry give way to free verse and then prose poetry in the past century. What has happened to AR? Is it still being used, or is it a historical curiosity? Is it still present but used less? Is it still being used but with a shift in focus to different forms of AR compared with the past? The *qaṣīda* has been historically dominant, but today there are many new genres of Arabic literature. What can we say about the use of AR in each of these genres? Which genres use AR more, and which use it less? What does all this mean for the future of AR?

Unfortunately, there are no data-driven answers to these questions. Fundamentally, there is no accurate[5] way to evaluate the level of AR in one text from a particular genre, time period or author, and compare it with the level of AR in another text. If we were able to do so, it would unlock the possibility of answering all the questions mentioned above, and others.

The remainder of this chapter will provide a novel conceptual framework for measuring the level of AR in Arabic texts, by measuring the density of AR LDs as a numeric marker for AR itself. Chapter 2 provides an explanation of how this conceptual framework was translated into the creation of the Arabic Rhetoric Literary Device Density Measurement System (ARLDDMS), a practical tool to measure the density of AR LDs in Arabic texts. Chapter 3 defines the LDs that the tool measures. Chapter 4 provides some results of real-life ARLDDMS scores that have been generated with this tool. Finally, Chapter 5 discusses the implications of this work, how it advances and makes a novel and unique contribution the field of AR, directions for further refinement of the tool and the research data it could generate.

[5] The meaning of “accurate” is explained in Section 2.1.

## 1.2) Is it possible to measure Arabic rhetoric?

The use of quantitative methods to analyse texts is well-established. Scoring systems to measure the lexical density of English texts have been devised by Ure[6] and Halliday[7]. Sophisticated tools to measure lexical density, grammatical density and grammatical complexity have found widespread application in the testing of computer-generated text, in academic testing, and to monitor changes in dialect use. Some of these tools have been adapted for Arabic,[8] and their range of applications is increasing.[9] The lexical density of English translations of Quranic verses has also been calculated.[10]

These tools have conceptually established the acceptability of counting the instances of the entity being measured (such as complex or foreign words, for example) in a text sample, and dividing by the number of words in the text sample, to give a density of the entity being measured. This density is numerically a representative proxy measure for the quality being measured, such as linguistic complexity or foreignness. It is not possible to measure linguistic complexity or foreignness directly, so the density of complex or foreign words is measured as a proxy.

It could be argued that the aesthetic, psychological and emotional effects of AR on the addressee cannot be measured with numbers. It is difficult to measure the emotional impact that a metaphor or simile evokes because there is tremendous variation between people. On the other hand, the "intelligence level" of a text can be assessed by measuring the lexical density, as previously mentioned. Pain, quality of life, sadness, severity of physical illness and likelihood of death can all be objectively measured in medical settings on well-validated and accepted numerical scales. Such scoring systems measure either the effect or the cause of the entity under investigation.

---

[6] Jean Ure, "Lexical density and register differentiation" in: G.E. Perren and J.L.M. Trim (Eds.), *Applications of linguistics: Selected papers of the Second International Congress of Applied Linguistics, Cambridge 1969* (Cambridge: University Press, 1971), 443-452.

[7] Michael A.K. Halliday, Christian.M.I.M. Matthiessen, *Halliday's Introduction to Functional Grammar* 4th Edition (New York: Routledge, 2014).

[8] Ahmed S. Al-Wahy, "Towards a Methodology for Measuring Lexical Density in Arabic" *International Journal of Arabic Linguistics* 3, 1, 1-33.

[9] Violetta Cavalli-Sforza, Hind Saddiki, Naoual Nassiri, "Arabic Readability Research: Current State and Future Directions" *Procedia Computer Science* 142 (2018), 38-49.

[10] Farhat Abdullah, Fatima Zahra, "Lexical Density and Cohesive Devices of the Divine Message in the English Translation of Surah Rehman" *International Bulletin of Literature and Linguistics* 3, 20 (2020), 38-50.

Similarly, the aesthetics of a painting are impossible to quantify numerically. Yet the integral components of what makes a good painting – such as the canvas, selecting the paints, the brushes, composition of the subject in relation to the foreground and background – can all be taught, learnt, improved and judged. The same applies to all art forms.

The tomes of AR books are irrefutable evidence that the Arabs have asked the question, “What makes for more effective, powerful and successful AR?” In doing so, they have analysed their speech and catalogued a wide range of features which contribute towards AR. This current project is a natural extension of this work, and asks the question, “Is it possible to measure the presence of these features numerically in a text?” While it would be impractical to attempt to measure the psychological effect of the Arabic rhetoric in the addressee, it could be possible to measure the density of AR LDs as a proxy.

A simple example will illustrate the fundamental basis of this premise. A famous celebrity, when asked in a media interview about whether they live in a big or small house, could describe their house by saying:

مساحة بيتي 200 متر مربع.

The area of my house is 200 square meters.

This sentence of six morphemes is entirely factual and contains no AR LDs. The same person could describe the same house in a different way:

بيتي كبير جداً، مثل قصر.

My house is very big, like a palace.

This sentence also has six morphemes and (i) replaces the probably unnecessary technical detail of the area of the house, and replaces it with an adjective “big” which is more contextually appropriate, and (ii) contains a metaphor by describing the house as being a “palace”. It therefore contains two AR LDs.

The same person could describe the same house in yet a different way:

يستحق الشخص الناجح إقامة التي تعكس مكانته الاجتماعية.

The successful person deserves a residence that reflects his social standing.

This sentence of 9 morphemes contains (i) an evasive response that forces the addressee to imagine the communicator’s house, (ii) a self-reference as a successful person (iii) ambiguity about whether the residence is a house or a flat or a palace, (iv) mention of an elevated social standing, (v) appropriateness to the context of a celebrity media interview, and (vi) a generalisable wisdom. This sentence therefore has six AR LDs.

Expressing the AR level as a density of LDs rather than as an absolute number of LDs will control for differences in word or morpheme count as the explanation for differences in numbers of LDs. The AR LD density counts of these three sentences are 0, 0.333 and 0.667 respectively. These three sentences demonstrate the face validity of measuring the AR LD density as a proxy measure of the level of AR in the text.

The aim of this study is to attempt to address the question, “Is it possible to objectively measure the density of AR LDs in Arabic texts, such that we can say that one text has a higher or lower LD density than another?” This project is designed to test the null hypothesis that it is not possible to objectively measure and compare the density of LDs in Arabic texts.

# Chapter 2
# Methodology

## 2.1) Introduction to the project methodology

The project was designed to create the ARLDDMS, a tool to accurately measure the density of AR LDs in a given sample to Arabic text. Accuracy in this context means that the tool would have:

- High sensitivity for detecting the presence of LDs. This means that if the LD is present, the tool would detect it.
- High specificity for detecting the presence of LDs. This means that if the LD is not present, the tool would not detect it.
- High inter-observer concordance in LD density scores. This means that if different assessors scored a text, they would detect the same number and density of LDs. An Arabic text under investigation should achieve the same LD density score irrespective of which assessor performs the scoring.
- Consistent scoring. This means that if the same assessor scored the same text more than once, the same number and density of LDs would be detected.

The ARLDDMS is a scoring system with which to count the number of AR LDs in a given sample of Arabic text, and by dividing this by the number of morphemes in the sample, to calculate the density of AR LDs in that sample of text.

Figure 1 summarises the project methodology.

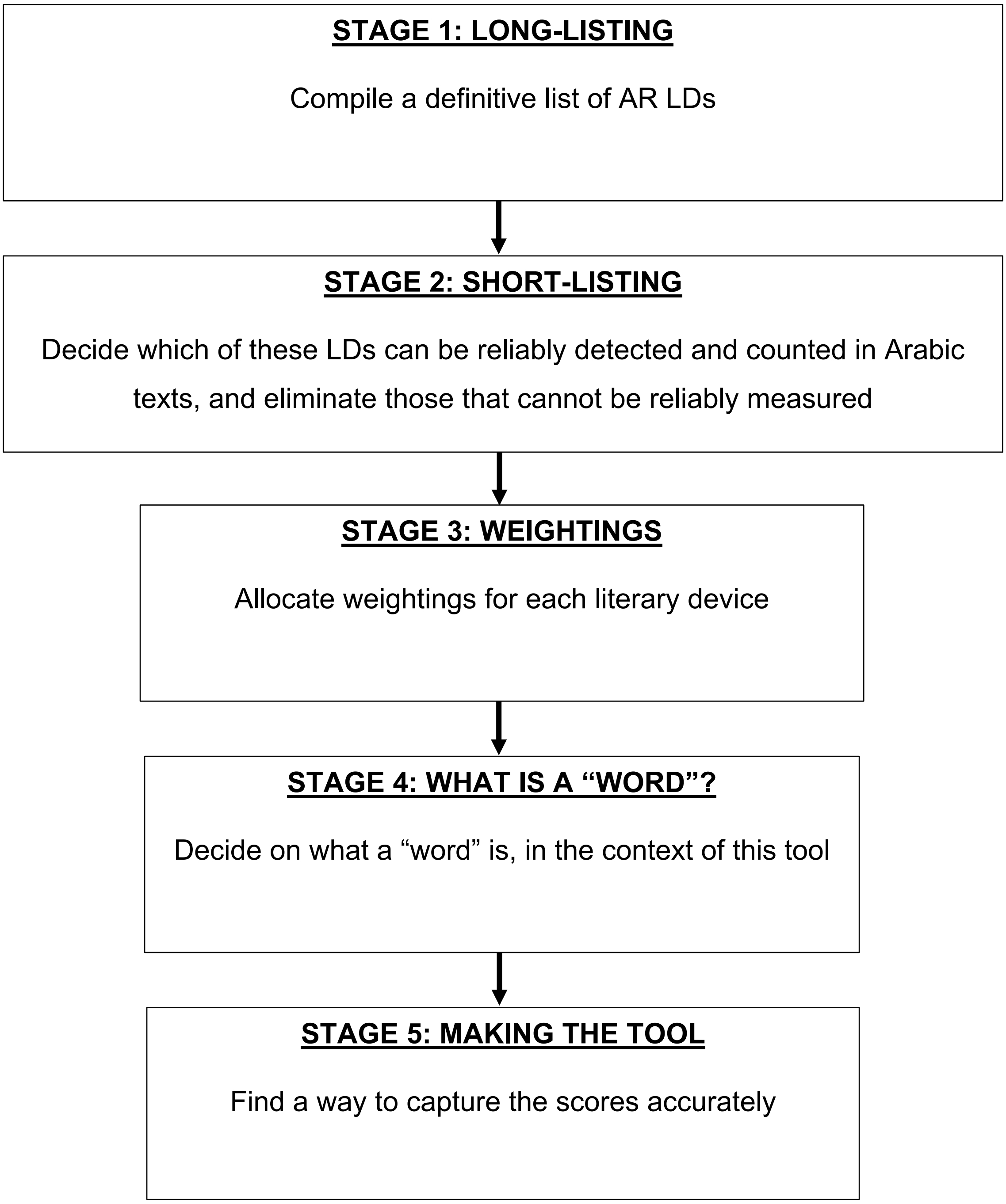

Figure 1: Summary of the project methodology

## 2.2) Stage 1: Long-listing of candidate Arabic rhetoric literary devices

The first step in making the tool was to compile a definitive list of all the AR LDs. This list was made as comprehensive as possible to enable the tool to have high sensitivity for AR LDs. The tool would explicitly state which LDs are included in the tool, and how to recognise them, as a way of ensuring high inter-observer concordance.

One of the major challenges for the project was that there is no universally accepted definitive list of AR LDs. This meant that the long list had to be created de novo.

Secondly, there is a profusion of books available on the subject of AR:

- A search on the website https://www.noor-book.com/ with search term "*balāgha*" (in Arabic) returned 523 book titles. A search with "balagha" (in English) returned 19 titles.
- Searches were also performed with search terms "*'ilm al-ma'ānī*", "*'ilm al-bayān*" and "*'ilm al-badī'*" (all in Arabic) similarly returned a large and impractical number of results.
- A search on https://books.google.com/ with the search term "balagha" in English identified one relevant book.
- A search on https://www.jstor.org/ with various search terms did not yield any useful resources.
- Discussion groups of non-native students of Arabic on social media were consulted regarding recommendations for suitable books for this project.
- Enquiries were made with personal contacts of experienced Arabic language teachers in Egypt.
- Textbooks used by the author in previous AR studies were reviewed.
- English and Arabic videos on AR were watched on https://www.youtube.com/

It was noted that most sources lean towards a particular genre of Arabic texts, such as poetry, prose or the Quran. Each genre has its own particular specialisms of LDs. For example, poetry-focused books mention details of poetic LDs related to rhyme and hemi-stich structure that are not relevant elsewhere.

Although the intention was to create a definitive list of literary devices, such a list would be far too long for the tool to be practically useful for Arabic texts in general. Additionally, the assessors would need to become familiarised with what would be minutiae within one genre. For example, some works of Quranic rhetoric run into over 11 volumes. It would be unrealistic to expect assessors to familiarise themselves with such detail. Ultimately, some of the sensitivity of the tool would have to be sacrificed in the interests of usability. Despite this, it is estimated at the real-world loss of sensitivity is likely to be small.

The result of this search process was that four books of AR[11] - two in Arabic, and two in English – were taken as the primary sources for this project. To mitigate against the bias that may be introduced by using sources that lean towards a particular genre of Arabic texts, the selection of primary sources reflects a spectrum of interests: the book by Abdul-Raof has a linguistics viewpoint, the books by Abd Al-Ghanī and Muhammad lean towards Quranic AR, while Nasif and colleagues draw on Arabic poetics.

A further challenge during compilation of the definitive list of AR LDs was the lack of standardisation and consistency in the identification, naming and definition of LDs. Different books refer to the same LD with different names and descriptions, and this is especially problematic when comparing English sources with Arabic sources. Some books describe a LD which another book mentions as a sub-type of another LD. This particularly happens with Quran-focused sources which have a higher granularity of LD definition than other books.

---

[11] H. Abdul-Raof, 2006.

Ayman Amin Abd Al-Ghanī, *Al-Kāfī Fī al-Balāgha: Al-Bayān, Wa al-Badiʿ Wa al-Maʿānī*, (Cairo: Dar Al-Tawfiqiyya Lil-Turath, 2011).

Hafnī Nāsif, Muḥammad Diab, Sultan Muḥammad, Muṣṭafā Tamum, *Durūs Al-Balāgha*, (Beirut: Dar ibn Ḥazm, 2012).

Hashim Muhammad, *First Steps to Understanding Balaghah*, (Leicester: Al-Qalam Publications, 2016)

A surprising finding was that different books dedicated markedly different amounts of space to describing the same LD. For example, assonance (*al-sajaʿ*) is covered in seven pages in one book[12] but only one small section in another.[13] Rhetorical shift (*al-iltifāt*) is covered in five pages in one book[14] but only 1 page in another.[15] Feigned ignornance (*tajāhul al-ʿārif*) is in the slimmest of the four sources[16] but not the others. One book details three levels of hyperbole[17] but others do not mention this detail.

All these discrepancies were concerning as they may have been symptomatic of underlying incoherencies in the importance attached to them by each author, how commonly each LD appears in Arabic texts, or how much the author is specialised in that domain. Ultimately it was decided that these differences should not detract from their role in the overall scoring system being created in the absence of any further evidence. The ultimate aim was to assemble all the relevant LDs irrespective of their name or categorisation. As previously mentioned, a balance had to be reached between having a high granularity and sensitivity of LD detection, versus the tool being practically usable.

To ensure that all the relevant AR LDs had been included in the long list, other sources were reviewed to see whether any additional LDs were found. No further LDs of significance were found. At the end of the long-listing process, 95 LDs were identified.

## 2.3) Stage 2: Short-listing of candidate Arabic rhetoric literary devices

An a priori list of criteria was formulated based on the objectives for accuracy described in Section 2.1 for LDs that would be removed from the long list:

- LDs which could be considered to be sub-types of other LDs such as "Caution" (*al-iḥtirās*) where the communicator provides additional information to prevent any misunderstanding. This function is already

[12] A. Abd Al-Ghanī, 2011, 193-199.
[13] H. Nasif et al, 2012, 119-120.
[14] A. Abd Al-Ghanī, 2011, 255-260.
[15] H. Abdul-Raof, 2006, 257-258.
[16] H. Muhammad, 2016, 48.
[17] H. Nasif et al, 2012, 112-113.

included under the category of “Verbosity” (*al-ʾiṭnāb*) and was therefore excluded.

- LDs which are more appropriately placed in other linguistic fields such as grammar or poetics rather than AR. Much of the material presented in books under the domain of word order and sentence construction (*ʿilm al-maʿānī)* is more appropriately considered under Arabic grammar rather than AR.
- LDs that are likely to have low inter-observer concordance. As an example, there are LDs that mention that a word is easy to pronounce for an Arab, or that the sentence flows smoothly, or that it has a musical flow.
- LDs where the intention of the communicator must be known as a pre-requisite for perceiving that the LD is present. Typically, the assessor will not know the communicator’s intention.
- Descriptions of some LDs rely on a judgement of whether the proposition as presented is the best way to say what the communicator intended. As the assessor does not know what the communicator intended, this makes this LD difficult to perceive, which would lead to a loss of consistency and inter-observer concordance.
- LDs which were essentially sub-categories or overlapping with others.

As a result of the short-listing process, the final list of AR LDs was 84. This process is summarised in Figure 2. Chapter 3 lists all the LDs included in the tool.

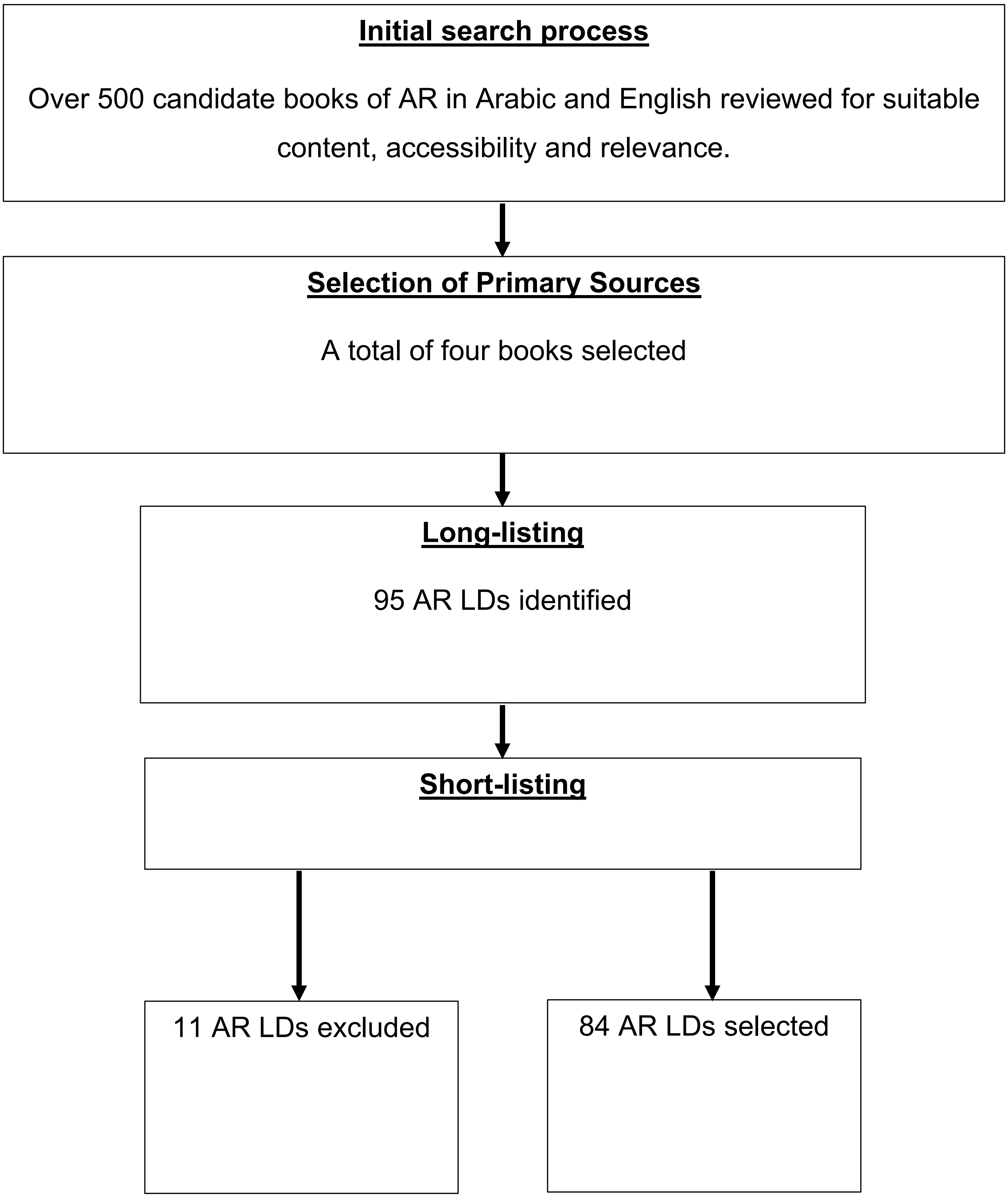

Figure 2: Summary of long-listing and short-listing process

## 2.4) Stage 3: Weightings and scoring system

Having decided which AR LDs should feature in the ARLDDMS, consideration was given to a system of weighting for each LD. The primary sources mention that within some LDs there are gradations of rhetorical strength. For example, an effective simile is a more intense form of AR than a detailed simile; a metaphor is a more intense form of AR than a simile. In assonance (*al-saja*ʿ), there are gradations based on the level of rhyme. It would therefore be reasonable to have a gradation of scores in such situations to reflect different levels of rhetoric.

Unfortunately, while such vertical gradations are mentioned, there is scant mention of corresponding horizontal gradations. For example, if a metaphor was awarded 2 marks whereas a simile was awarded only 1 mark, this would incorrectly indicate that a metaphor was twice the rhetorical value of other devices apart from simile. Due to this, it was decided to not implement a system of weightings for any literary device.

The one exception is the very final LD in the ARLDDMS, which is not an LD in itself. Rather, it is a collection of linguistic features that detract from rhetoric value of a text. A mark will be deduced from the overall score for each occurrence of the defects mentioned in that section.

Consideration was given to whether the occurrence of each LD in a text should score one mark, or whether the marking system should range from zero to ten or even one hundred. Having a binary system with zero (literary device not present) and one (literary device present) does not allow the assessor to express any opinion on the excellence of the execution of that literary device, which would be a wasted opportunity.

On the other hand, allowing assessors the discretion to award up to ten marks for an occurrence of a LD runs the risk of a very wide variation in the spread of marks for a given Arabic text, which would reduce inter-observer concordance. Generous assessors may award high marks; less generous assessors may consistently award lower marks. This variation becomes a problem when the marks for two texts are subsequently compared: the distributions of marks overlaps, which means that it is not possible to discern a true difference between the texts, resulting in a type 2 measurement error. This difference is visually represented in Figure 3, which is based on modelling data shown in Appendix 4.

An analysis of the instances where an assessor is tempted to award exceptionally high scores for one lexical item was performed. It was found that typically this occurred because the lexical item incorporated more than one literary device, such as a metaphor and an onomatopoeia and a personification appearing together. It would be more accurate for assessors to award marks for each LD in a given lexical item, such as 1 mark each for metaphor, onomatopoeia and personification, for example. As an another example, a pun may score under pun, metonym, or metaphor which are all types of figurative speech. The assessor should award a score only once for this. However, a pun that is also an onomatopoeia would score for the figurative speech as well as for the onomatopoeia.

Ultimately, it was decided to allow assessors to award 0, 1 or 2 marks for each occurrence of an LD.

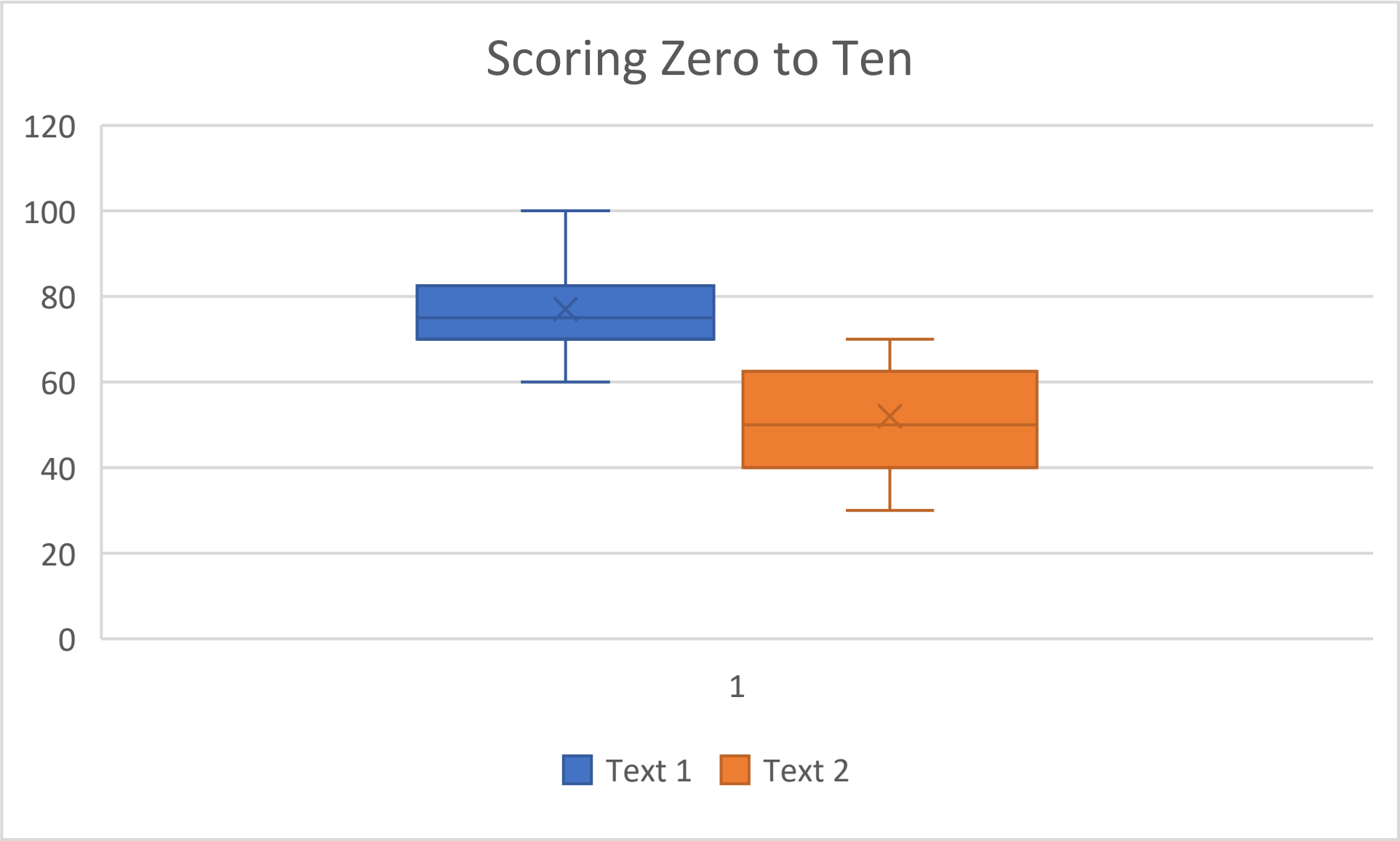

In this scenario, Text 1 has 10 LDs and Text 2 has only 5 LDs. Assessors may score from 0 to 10 for each LD. This results in a wide range of scores. The real difference between Text 1 and Text 2 cannot be detected as the ranges of scores overlap.

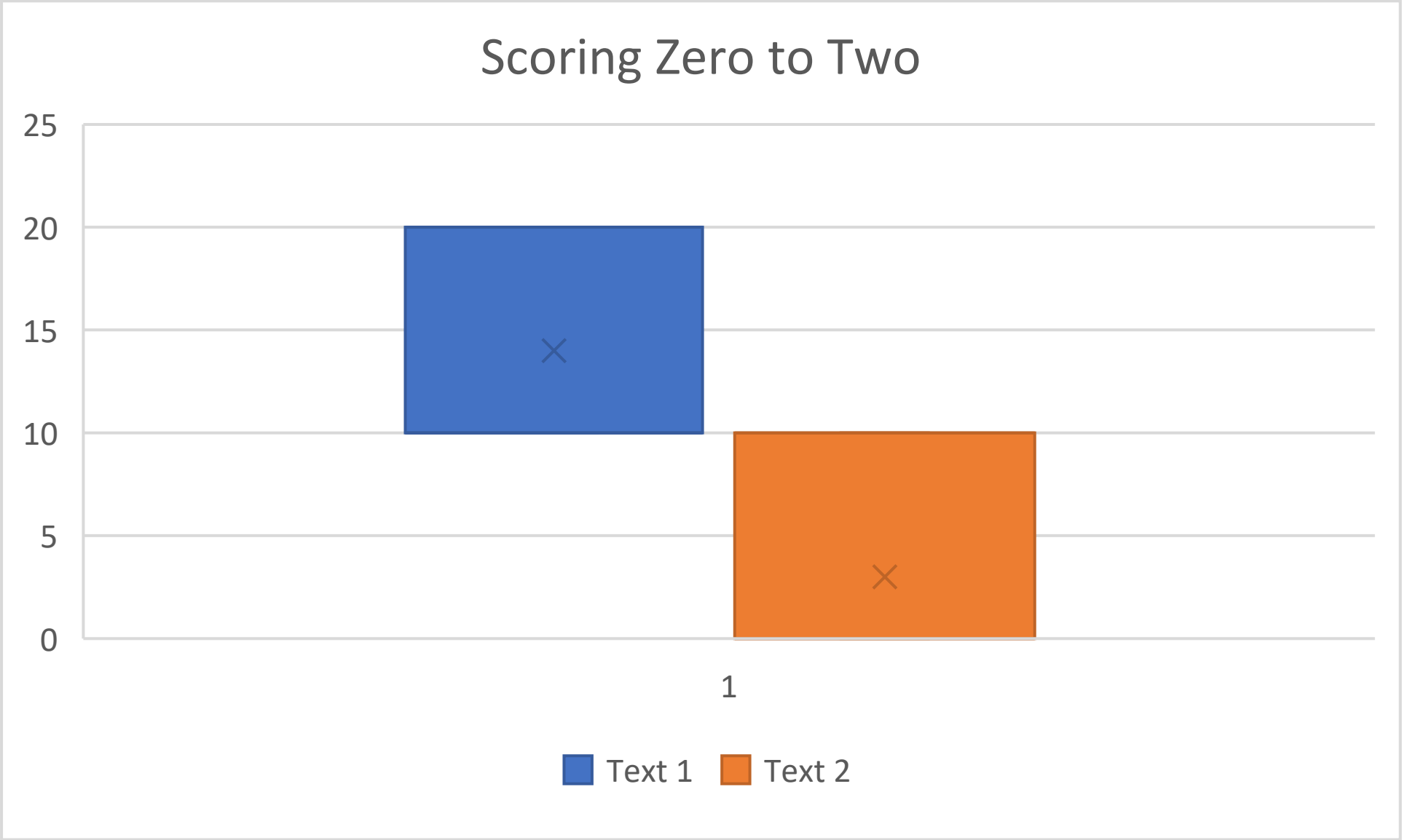

In this scenario, Text 1 still has 10 LDs and Text 2 still has only 5 LDs. Assessors may now score from 0 to 2 for each LD. This results in a narrower range of scores. The real difference between Text 1 and Text 2 is now apparent.

Figure 3: Illustration of the difference in sensitivity of the tool when the range of awardable scores is changed

## 2.5) Stage 4: What is a "word" in Arabic?

When counting the number of words for calculating the AR LD density, it must be borne in mind that Arabic is a highly synthetic language which relies on the inflection and agglutination of morphemes. For example, this sentence has six words, but ten morphemes:

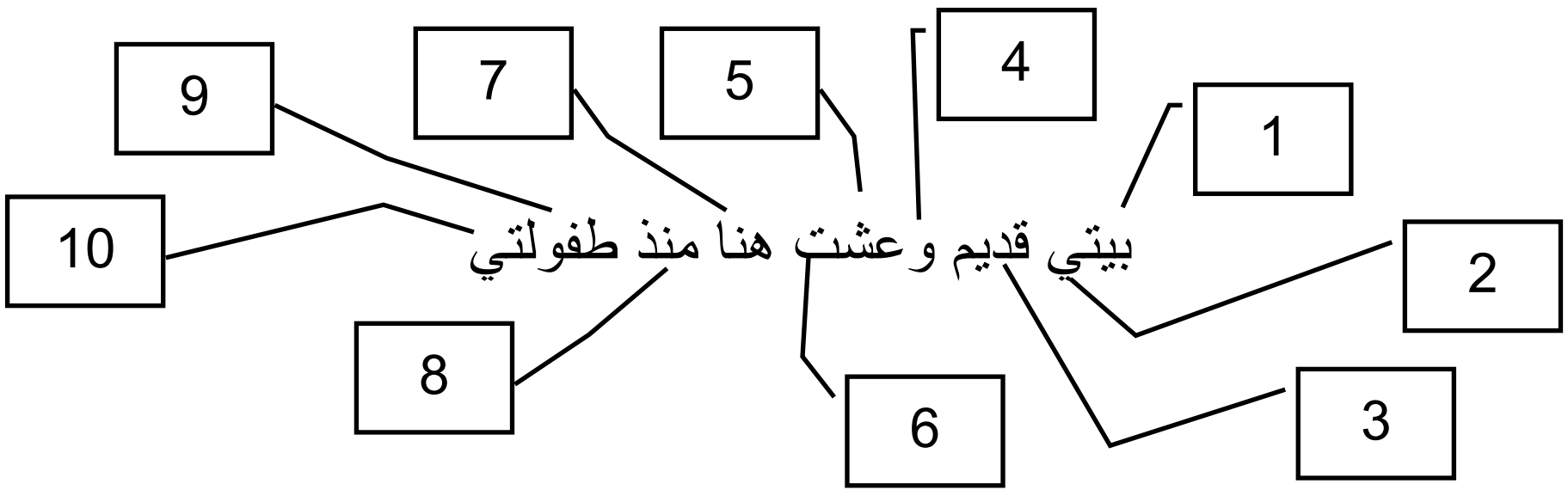

My house is old and I have lived here since my childhood

It is therefore more appropriate to count the number of morphemes rather than the number of words. This is consistent with previous work in this field.[18] The definite article *al-* is considered a marker of definiteness and not a distinct morpheme for the purpose of this tool.

## 2.6) Stage 5: The practicalities of creating a usable tool

Chapter 3 lists all the 84 LDs included in the tool. For ease of reference, they are divided into three major domains as per the modern division of AR into Word Order and Sentence Structure, Figurative Speech, and Embellishments. Each LD is introduced with a reference code signifying the domain it is included under, and its name in both English and Arabic to help familiarise assessors. This is followed by a definition of the LD, for the purpose of this tool. Examples have been provided in Arabic and English where useful to do so. Where a LD may have various sub-types, a brief summary of these is presented to help identify occurrences in the texts. A reference is provided to the primary source from which each LD is taken, so that assessors may clarify their understanding of the definition if needed. A reminder of the possible scores per occurrence of the device in the text is provided. This chapter is not intended to be a text-book or primer of AR LDs. It is envisaged that assessors will already be familiar with the field, but these definitions serve to ensure that all assessors have the same understanding. They serve as an aide-memoire and a final arbiter in case of any disagreement.

[18] A.S. Al-Wahy.

It is evident that different assessors may consider a particular lexical item to be classifiable under more than one device. For example, a metonymy may also be a simile or metaphor,[19] or a head-tail may also be a *jinās*. Some of the primary sources mentioned that there are scholarly differences on the classification of certain literary devices.[20] As mentioned previously in the context of multiple LDs occurring together, the assessor must recognise the occurrence as one instance of LD and award a mark as such, irrespective of the title or classification given to it.

Once an assessor has identified and marked all the LDs in a text sample, the total number of LDs is calculated by summing all the scores from each domain. The assessor must then count the number of morphemes in the text sample. The LD density is calculated by dividing the sum total of LD by the number of morphemes.

A peculiarity of such a scoring system is that it is conceivably possible for a small extract of text to consist of 3-5 morphemes but contain a high number of LDs. Such a text sample would have a high LD density, but not be representative of the author, genre or epoch. Therefore the ARLDDMS was designed to not report only the raw LD density, but a representation of the spread of LDs across the three domains, and a representation of the morpheme count which has generated the density score. This is represented by the Arabic Rhetoric Literary Device Density Summary where the breakdown of scores according to domain is visually presented as the numerator, and the morpheme count as the denominator:

[19] A. Abd Al-Ghanī, 2011, 97-98.
[20] A. Abd Al-Ghanī, 2011, 127.

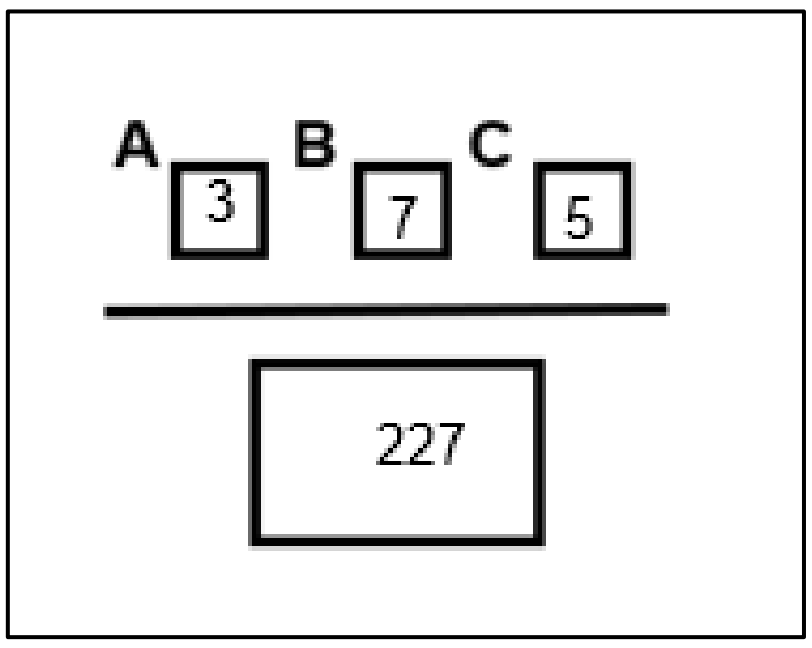

Representing the tool's output in this way also gives useful information about the distribution of LDs across the three domains in a given text.

In Domain C (Embellishments), it was noted that the primary sources consistently classify LDs into semantical embellishments (*al-muḥsināt al-maʿanawiyya*) and lexical embellishments (*al-muḥsināt al-lafẓiyya*), respecting the traditional classification of this field. However, this resulted in an unwieldy and unhelpful collection of approximately 20 devices in the former and 40 devices in the latter. This traditional classification was not granular enough for use in this tool.

The LDs in this domain were therefore re-categorised into seven new categories. They were categorised not according the manner in which they functioned (semantic vs lexical) but rather according to where in the text the assessor might locate them. These categories start at the word-unit level, and progress to the level of the whole text. The new classification was more useful for the purpose of this tool: all the related LDs were positioned close to each other, so it became easy to decide which LD was present in the text. This increased inter-observer concordance in scoring.

To help assessors accurately score texts, a data collection pro-forma was created (Appendix 1) which brings together the 84 LDs, their scores, the morpheme count and the final ARLDDMS score. With the names of 84 LDs written on 4 pages of paper, it can be difficult to remember what each of the LDs is. This can lead to errors in scoring. To mitigate against this, Quick Response (QR) matrix barcodes were designed and placed on the proforma. They were designed so that if an assessor wishes to know further information about a particular LD's definition, they can scan the QR code with their mobile device, and this will load a website where the LD's definition is provided.

An internet domain name (https://www.balaghascore.com/) was registered and a website created to drive this functionality, and to support the project in general. To provide a more user-friendly means of using the ARLDDMS tool, an online calculator was created on the website, accessible at https://balaghascore.com/online. Assessors may access this online calculator from any internet-connected mobile device or desktop computer anywhere in the world at any time of day or night. It provides a means to input the scores for each LD in a text, along with links to the literary device definitions the project is using. When the morpheme count is inputted, the calculator presents the ARLDDMS score and Arabic Rhetoric Literary Device Density Summary instantaneously. Appendix 2 provides screenshots of this.

The tool was ported to a Microsoft Excel spreadsheet (see Appendix 3) which may be downloaded from https://balaghascore.com/Arabic-Rhetoric-Density-Calculator.xls and an Adobe PDF document which may be downloaded from https://balaghascore.com/Arabic-Rhetoric-Scoring-Proforma.pdf for offline use.

# Chapter 3

# The Arabic Rhetoric Literary Device Density Score

## 3.1) Domain A: Word Order and Sentence Structure

The rules of Arabic grammar dictate the normal syntax of Arabic constructions. In order to award marks for an LD in this domain, the communicator must depart from the expected grammatical syntax in order to exert a rhetoric effect.[21]

| **A-1: Use of a reporting sentence where an informing sentence is expected, and vice versa[22]** | **ظهر جملة خبرية في مكان مقتضى لجملة إنشائية وعكسه** |
|---|---|
| ***Definition***<br>Lexical items may be classified into two:<br>1) Reporting sentences which convey factual information such as:<br>البيت كبير.<br>The house is big.<br>2) Informing sentences which do not convey factual information such as:<br>هل البيت كبير؟<br>Is the house big?<br>Informing sentences may be used to express the interrogative, the imperative, prohibition, the vocative and wishes, praise, dispraise, astonishment, hope, oaths and legal contracts.<br>A reporting sentence may be used when an informing sentence is expected, and vice versa, for rhetorical purposes. | |

[21] Muhammad Abdel Haleem, "Rhetorical Devices and Stylistic Features of Qur'anic Grammar" in: Mustafa Shah and Muhammad Abdel Haleem (Eds.) *The Oxford Handbook of Qur'anic Studies* (University Press: Oxford, 2020), 333.

[22] H. Abdul-Raof, 2006, 103-121 & H. Muhammad, 2016, 18-19.

***Examples***

1) Use of an informing sentence in place of an expected reported sentence:

فاطمة: هل أكلتَ التفاحة؟

سليم: ألا تعلمين أنني آكل تفاحة كل صباح؟

Fatima: Did you eat the apple?

Salim: Don't you know that I eat an apple every morning?

2) Use of a reporting sentence in place of an expected informing sentence: When caught cheating in an exam, a student expects a severe rebuke from the teacher in the form of an informing sentence such as "Never try to cheat again, or you'll be expelled!" Instead, the teacher delivers a reporting sentence:

المعلم: الغش انتهاك جسيم للثقة المقدسة بين المعلم وطالبه.

Teacher: Cheating is a severe violation of the sacred trust between the teacher his student.

***Scoring criterion***

Marks may be awarded if there is a deviation from the expected norms of reporting and informing sentences, for the creation of a rhetorical effect.

| | | |
|---|---|---|
| **Score** | This literary device is not present | **0** |
| | This literary device is present | **1** |
| | This literary device is present, with the highest rhetorical effect | **2** |

| A-2: Affirmation[23] | التأكيد |
|---|---|

***Definition***

Affirmation of the communicator’s proposition which is being conveyed in a reporting sentence takes place through the use of certain affirmatory constructions such as:

- The particle ʾ*inna* starting a nominal sentence.
- The particles *la*, ʾ*in*, ʾ*an*, *mā*, *lā*, and *min*, and the prefix *bi*-.
- The construction ʾ*ammā… fa*-.
- The future-tense prefix *sa*- and the future-tense particle *sawfa*.
- The use of *qad* before a past-tense verb.
- Explicit pronouns.
- Oath letters such as *wa*-, *bi*- or *ta*- prefixed to another oath word such as “Allah”.
- Light affirmation with the suffix -*n* and heavy affirmation with the suffix -*nna*.
- Repetition of a negation.
- The particle ʾ*innamā*.
- Affirmative words such as *ḥattā*.
- Use of the negation followed by an exception.

Depending on the status of the addressee, reporting sentences may utilise 3 levels of affirmation:

1) No affirmation: The addressee is open-minded or neutral about the communicator’s proposition: there is no requirement for affirmation.

سليم: أنا مريض.

Salim: I am ill.

[23] A. Abd Al-Ghanī, 2011, 401-407 & H. Abdul-Raof, 2006, 108-114.

2) Light affirmation: The addressee is doubtful or sceptical about the communicator’s proposition, and there is a requirement for affirmation by using one affirmatory device.

سليم: إنَّني مريض.

Salim: Verily I am ill.

3) Heavy affirmation: The addressee is in denial about the communicator’s proposition: there is a requirement for more intense affirmation by using two or more affirmatory devices.

سليم: إنَّني لمريض.

Salim: Verily I am truly ill.

The affirmative may be used for in a way outside this scheme of escalation, for rhetorical purposes.

***Examples***

- An open-minded or neutral addressee is addressed as if they are doubtful, with one affirmatory device.

كريم: أنت جاهز للسفر؟

سليم: لا أريد أن أسافر اليوم. إنَّني مريض.

Karim: Are you ready to travel?

Salim: I do not want to travel today. Verily, I am ill.

- An open-minded or neutral addressee is addresses as if they were in denial, with two or more affirmatory devices.

كريم: أنت جاهز للسفر؟

سليم: لا أريد أن أسافر اليوم. إنَّني لمريض فلن أخرج من بيتي إلا عند الضرورة.

Karim: Are you ready to travel?

Salim: I do not want to travel today. Verily I am truly ill,

| |
|---|
| and so I will not leave my house unless it was necessary.<br><br>• An addressee in denial is addressed as if they were open-minded or neutral, without any affirmatory devices.<br>كريم: كلنا جاهزون للسفر والسيارة جاهزة أيضاً. أين أنت؟ ننتظر إليك...<br>سليم: نصحني الطبيب بالبقاء بالبيت اليوم.<br><br>Karim: We are all ready to travel, and the car is ready too.<br>Where are you? We are waiting for you.<br>Salim: The doctor advised me to stay home today.<br><br>***<u>Scoring criterion</u>***<br>No mark is awarded for the use of an affirmatory device per-se. Marks may be awarded only if there is a deviation from the expected norms of affirmation, for the creation of a rhetorical effect. |

| | | |
|---|---|---|
| **Score** | This literary device is not present, or is not used in a manner that deviates from the norm | **0** |
| | This literary device is present | **1** |
| | This literary device is present, with the highest rhetorical effect | **2** |

| **A-3: The Imperative**[24] | **الأمر** |
|---|---|
| ***<u>Definition</u>***<br>The role of the imperative is to communicate a compelling command from someone in a position of authority to a subordinate. The imperative is may be recognised by the following:<br><br>• A verb in the imperative case | |

[24] H. Abdul-Raof, 2006, 117-118.
A. Abd Al-Ghanī, 2011, 331-335.
H. Nasif et al, 2012, 31-32.
H. Muhammad, 2016, 20-22.

- Prefix of *li* to a present tense verb in the jussive mood
- Use of ʿ*alā* and *hattā*
- Use of verbal noun in place of the imperative

The imperative may be used for purposes other commanding, for rhetorical purposes.

***<u>Examples</u>***

1) To advise:

أُدرسْ جيدًا لهذه الاختبارات!

Study well for these exams!

2) To give a choice:

خُذْ هذا الكتاب أو تلك المجلة.

Take this book or that magazine.

3) To denote an equal choice:

زرني أو لا تزرني اليوم، سأراك غداً في أي حال.

Whether or not you visit me today, I will see you tomorrow in any case.

4) To supplicate:

ربِّ زِدْنِي عِلْماً

Quran 20:114

“Oh my Lord, increase me in knowledge!”

5) To challenge:

عِشْ حياتك بدون خوف من أي بشر.

Live your life without fearing any human.

6) To threaten:

أُحضرْ إلى الشغل متأخراً مرة ثانية والمدير سيطردك.

Come to work late again, and the manager will fire you.

7) To express sarcasm:

اِفعلْ ما تشاء، لا يهمني.

Do whatever you want, I don't care.

8) To plea:

ساعدني!

Help me!

9) To give permission:

خذ راحتك.

Take your time.

10) For a weaker person to petition someone in authority:

أيها الملك، كن كريماً تجاه شعبي.

Oh King, be generous towards my people!

11) To insult:

كن كسولاً مثل أصدقائك.

Be lazy like your friends.

***<u>Scoring criterion</u>***

No mark is awarded for the use of the imperative to issue a command. Marks may be awarded only if there is a deviation from the expected norms of the imperative, for the creation of a rhetorical effect.

| | | |
|---|---|---|
| **Score** | This literary device is not present, or the imperative is used to merely issue a command | **0** |
| | This literary device is present | **1** |
| | This literary device is present, with the highest rhetorical effect | **2** |

| A-4: Prohibition[25] | النهي |
|---|---|

***Definition***

The role of the prohibitive is to communicate a compelling prohibition. The prohibitive may be recognised by the use of *lā* and the present tense in the jussive mood. The prohibitive may be used for purposes other than prohibition, for rhetorical purposes.

***Examples***

1) To supplicate:

رَبَّنَا لَا تُؤَاخِذْنَآ إِن نَّسِينَآ أَوْ أَخْطَأْنَا

Quran 2: 286

“Our Lord! Do not punish us if we forget or make a mistake.

2) To threaten:

لا تستخدم حزام الأمان.

Do not use a seatbelt.

(Implied threat of harm in the event of an accident)

3) To express sarcasm:

لا تمارس الرياضة لأن صحتك ليست مهمة.

Don’t do any exercise, because your health is not important.

4) To guide:

لا تضيع وقتك

Do not waste your time

5) To challenge:

لا تحاولْ الحصول على درجات أفضل من أخيك.

Do not try to get better grades than your brother.

[25] H. Abdul-Raof, 2006, 118-119.
A. Abd Al-Ghanī, 2011, 336-339.
H. Nasif et al, 2012, 32-33.

6) To rebuke:

لا تفخرْ كثيراً بإنجازاتك.

Do not be so proud of your achievements.

7) To express a desire:

يا نقودي، ابق في جيبي

Oh money, stay in my pocket!

***Scoring criterion***

No mark is awarded for the use of the prohibitive to issue a simple prohibition. Marks may be awarded only if there is a deviation from the expected norms of the prohibitive, for the creation of a rhetorical effect.

| | | |
|---|---|---|
| **Score** | This literary device is not present, or the prohibitive is used to merely prohibit | **0** |
| | This literary device is present | **1** |
| | This literary device is present, with the highest rhetorical effect | **2** |

| **A-5: The Interrogative**[26] | **الاستفهام** |
|---|---|

***Definition***

The interrogative is used to ask a question. A number of interrogative participles may be used, including hamza, *ʾalam*, *ʾalan*, *matā*, *ʾayna*, *kayfa*, *kam*, *mā*, *ʾannā*, *hal*, *ʾayy*, *man*, and *ʾayyāna*. The interrogative may be used for purposes other than asking questions, for rhetorical purposes.

***Examples***

1) To command:

هل ستنظف غرفتك؟

[26] H. Abdul-Raof, 2006, 115-117 & 256.
H. Nasif et al, 2012, 33-38
A. Abd Al-Ghanī, 2011, 340-352.

Will you clean your room? (Which means – clean your room!)

2) To prohibit:

هل تخاطر بصحتك بالتدخين؟

Do you take a risk with your health by smoking?

3) To rebuke:

كيف تعرف أنه لن يكتشف أحد ما فعلته؟

How do you know that no one will find out what you did?

4) To express sarcasm, or astonishment:

ما هذا المكان؟

What is this place?

5) To warn:

من يتذكر ما حدث آخر مرة اشتكينا فيها؟

Who remembers what happened the last time we complained?

6) To threaten:

هل ترغب مني أن أغضب؟

Do you want me to become angry?

7) To deny:

هل اشتريت هذا المنزل الكبير بنفسك؟

Did you buy this house yourself?

8) To express an impossibility:

هل تظن أنه يمكنك الهروب من قدرك؟

Do you think you can escape from your destiny?

9) To express pride:

هل تعرف اي قبيلة أنتمي؟

Do you know which tribe I belong to?

10) To ask a rhetorical question that is intended to produce an effect rather than an answer:

هل رأيت سعر البنزين اليوم؟

Have you seen the price of petrol today?

***Scoring criterion***

No mark is awarded for the use of the interrogative to ask a simple question. Marks may be awarded only if there is a deviation from the expected norms of the interrogative, for the creation of a rhetorical effect.

| | | |
|---|---|---|
| **Score** | This literary device is not present, or the interrogative is used to merely ask a question | **0** |
| | This literary device is present | **1** |
| | This literary device is present, with the highest rhetorical effect | **2** |

| **A-6: Wish[27]** | **التمنى** |
|---|---|

***Definition***

The particles *laʿlla* and *ʿasā* are used in the context of a wish that an outcome which is possible would occur. The particles *layta*, *law* and *hal* are used to wish that an impossible outcome may occur. The use of particles normally used with possible things but to denote impossibility, and vice-versa, can have rhetorical effects.

***Example***

أمارس الرياضة لساعة واحدة كل يوم، لعّلي أمنع اقتراب الشيخوخة به.

I exercise for an hour every day, hoping to prevent the approach of old age by it.

---

[27] H. Abdul-Raof, 2006, 119-120.
A. Abd Al-Ghanī, 2011, 353-356.
H. Nasif et al, 2012, 38

| **Scoring criterion** | | |
|---|---|---|
| No marks are awarded for using these particles in their expected usage. Marks may be awarded when these particles are used in other ways to achieve rhetorical effects. | | |
| **Score** | This literary device is not present | **0** |
| | This literary device is present | **1** |
| | This literary device is present, with the highest rhetorical effect | **2** |

| **A-7: The Vocative[28]** | **النداء** |
|---|---|

***Definition***

The vocative is denoted by the use of hamza or ʾ*ayy* when the addressee is near, and *yā*, ʾ*ā*, ʾ*āy*, ʾ*yā*, and *hayyā* for an addressee that is far from the speaker. The particle *wā* is used to address a mourner. These conventions may be broken for rhetorical effect: close-by addressees may be addressed as if they are far away, and vice-versa, to denote physical or emotional distance from the communicator. The vocative may also be used to introduce the expression of regret, lamentation, sarcasm or rebuke.

***Example***

هيّا سليم، انتبه!

Oh Salim, pay attention!

***Explanation***

Salim is close to the communicator but has been addressed as if he was far away, to highlight that Salim is not paying attention and is psychologically far away from the communicator.

[28] H. Abdul-Raof, 2006, 119.
A. Abd Al-Ghanī, 2011, 357-363.
H. Nasif et al, 2012, 38-40.

| |
|---|
| ***Scoring criterion***<br>No marks are awarded for using vocative particles for their expected usage. Marks may be awarded when vocative particles are used in other ways to achieve rhetorical effects. |

| | | |
|---|---|---|
| **Score** | This literary device is not present | **0** |
| | This literary device is present | **1** |
| | This literary device is present, with the highest rhetorical effect | **2** |

| **A-8: Definiteness and Indefiniteness**[29] | **التعريف والتنكير** |
|---|---|

***Definition***

Definiteness and indefiniteness can be marked by the use of proper nouns, nicknames, personal pronouns, demonstrative pronouns, relative pronouns, genitive constructions, the vocative and the definite article al-. Rhetorical effects may be achieved by the use of definiteness where indefiniteness is expected, and vice-versa, as well as by their use in unexpected ways.

***Examples***

1) Use of demonstrative pronouns to glorify or belittle:

أهذا مدير الشركة؟

Is *this* the manager of the company?

In this example, the use of *hadha* (this) rather than *huwa* (he) is a way of belittling the person.

2) Use of relative pronoun to conceal someone's identity:

قابلتُ الذي كتب لي الرسالة الغريبة.

I met the one who wrote me the strange letter.

[29] H. Nasif et al, 2012,49-56 & H. Muhammad, 2016, 29-36.

The communicator has used *alladhī* (who) to avoid mentioning the name of the person.

***Scoring criterion***

Marks may be awarded when the lexical item demonstrates the use of definiteness and indefiniteness out with the normal grammatical rules governing their use, for a rhetorical effect.

| | | |
|---|---|---|
| **Score** | This literary device is not present | **0** |
| | This literary device is present | **1** |
| | This literary device is present, with the highest rhetorical effect | **2** |

| **A-9: Length of Speech: Brevity, Verbosity and Moderation**[30] | الإيجاز والإطناب والمساواة |
|---|---|

***Definition***

The lexical item under investigation should demonstrate a length which is appropriate to the context and the communicative needs demanded of it. Brevity and succinctness may be achieved by:

1) using precise words which convey expansive meanings and
2) omission of lexical items which are not essential to conveying the intended meaning, which could be inferred, or which are already known to the addressee.

Conversely, it may be appropriate to be verbose in order to achieve a rhetorical outcome. This may occur:

1) when more information, details or clarification are required to be transmitted,
2) when further clarification is required,
3) As a means of affirmation of the information already supplied,

[30] H. Abdul-Raof, 2006, 188-192.
A. Abd Al-Ghanī, 2011, 381-397.
H. Muhammad, 2016, 42-46.

4) To influence the addressee by appealing to the addressee's emotions.

A text is said to show Moderation when there is a balance between verbosity and brevity.

***Scoring criterion***

Marks may be awarded if there is evidence of brevity, moderation or verbosity being used to create a rhetorical effect.

| | | |
|---|---|---|
| **Score** | There is no evidence that brevity, moderation or verbosity have been sought | **0** |
| | Brevity, moderation or verbosity are present | **1** |
| | Brevity, moderation or verbosity are present, giving the highest rhetorical effect | **2** |

| **A-10: Foregrounding and Backgrounding / Anastrophe[31]** | **التقديم والتأخير** |
|---|---|

***Definition***

Foregrounding is the bringing forwards of particular lexical items to nearer the beginning of the sentence. This emphasises or gives more importance to the lexical items which are foregrounded. This can be utilised in the following situations:

1) To convey implied information about the hierarchy of the lexical items within the sentence.
2) To give good news or reassurance before giving bad news or criticism
3) To control the order in which information is given, such as starting with the general, then proceeding to the specific, or vice-versa.
4) To express astonishment, or to arouse fear in the addressee

[31] H. Abdul-Raof, 2006, 121-122.
H. Nasif et al, 2012, 45-47.

Conversely, backgrounding is the moving of lexical items towards the end of the sentence. It may reduce the emphasis on those items, or it may also build a sense of suspense in the addressee.
Foregrounding and backgrounding may also be employed for the purpose of the rhythm or metre of a sentence or verse of poetry.

***Scoring criterion***

Marks may be awarded if there is evidence of foregrounding or backgrounding to achieve a rhetorical effect.

| | | |
|---|---|---|
| **Score** | This literary device is not present | **0** |
| | This literary device is present | **1** |
| | This literary device is present, with the highest rhetorical effect | **2** |

| **A-11: Ellipsis[32]** | الذكر والحذف |
|---|---|

***Definition***

The omission of certain lexical items to create a rhetorical effect, such as:

1) Making the sentence more succinct
2) Where the omitted information is already known to the addressee
3) To convey emotion more succinctly
4) To evoke an immediate response from the addressee
5) For the sake of rhyme, metre or musicality
6) Evoking suspicion or interest by concealing the identity of the subject
7) To convey some fixed expressions

***Scoring criterion***

Marks may be awarded if it is discernible that ellipsis has been employed to achieve a rhetorical effect.

[32] H. Abdul-Raof, 2006, 133-136 and 159-160.
H. Nasif et al, 2012, 41-44.

| Score | | |
|---|---|---|
| | This literary device is not present | **0** |
| | This literary device is present | **1** |
| | This literary device is present, with the highest rhetorical effect | **2** |

| A-12: Exophora[33] | الإضمار مقام الإظهار |
|---|---|

***Definition***

A pronoun is used without defining it, because the meaning is understood from the context, or due to the addressee's previous knowledge.

***Example***

When Salim returns from having met Karim:

فاطمة: ماذا قلت له؟

Fatima: What did you say to him?

| Score | | |
|---|---|---|
| | This literary device is not present | **0** |
| | This literary device is present | **1** |
| | This literary device is present, with the highest rhetorical effect | **2** |

| A-13: Use of Noun in Place of Pronoun[34] | الإظهار مقام الإضمار |
|---|---|

***Definition***

The communicator does not use a pronoun even though the proper name has already been mentioned. This emphasises and highlights the proper name.

***Example***

When Salim returns from having met Karim:

فاطمة: ماذا قلتَ لكريم؟ قل لي ماذا قلتَ لكريم؟ أريد أن أعرف ما قلتَ لكريم.

Fatima: What did you say to Karim? Tell me what you said to Karim! I want to know what you said to Karim.

33 H. Muhammad, 2016, 47.

34 H. Muhammad, 2016, 47-48.

| Score | | |
|---|---|---|
| Score | This literary device is not present | **0** |
| | This literary device is present | **1** |
| | This literary device is present, with the highest rhetorical effect | **2** |

| **A-14: Use of Appropriate Linguistic Style and Register**[35] | استخدام الأسلوب المناسب |
|---|---|

***Definition***

The communicator uses a linguistic style and register which is appropriate to the subject matter and the intended audience. For example:

- a scientific article would use relevant technical jargon and an academic style of writing to convey factual information,
- a work of fiction such as a novel may use imagery and figurative language to evoke an emotional response in the reader,
- a non-fiction work such as a book of national history may mix factual and non-factual stylistic elements to convince the reader.

| | | |
|---|---|---|
| **Score** | Inappropriate linguistic style or register | **0** |
| | Somewhat appropriate style and register, but not entirely | **1** |
| | Totally appropriate linguistic style and register | **2** |

## 3.2) Domain B: Figures of Speech

| **B-1: Simile**[36] | التشبيه |
|---|---|

***Definition***

A simile is a literary device where something – termed "the likened-to" (*al-mushabbah*) – is described as being compared with something else – termed "the likened" (al-*musbabbah bihi*) with which it shares a common attribute – termed "the simile feature" (*wajah al-tashbīh*). The likened is more intense in the simile feature than the likened-to. The two ends of the simile (*al-ṭarafān*) – the likened-to and the likened – are connected by the simile element (*ʾadāt al-tashbīh)* which may be:

- The particles *ka-* (“as”) and *kaʾanna* (“as if”)
- Verbs like *shābaha*, *māthala*, *ḍāhā*, *ḥākā*, *ḍāraʿa* connotating resemblance and similarity
- Nouns such as *mathala*, *mathīl*, *shibh*, *shabīh*, *mushābih*, *muḍāriʿ* connotating resemblance and similarity.

[35] A. Abd Al-Ghanī, 2011, 414-418.
[36] H. Abdul-Raof, 2006, 198-209.
A. Abd Al-Ghanī, 2011, 42-64.

Various components of the simile can be omitted to create a rhetorical effect:

| | |
|---|---|
| Detailed simile (*al-tashbīh al-mufaṣṣal*) – all four elements are present. | سليم كأسد في الشجاعة.<br>Salim is like a lion in courage. |
| Single simile (*al-tashbīh al-mufrad*) – the simile feature is omitted. | سليم كأسد.<br>Salim is like a lion. |
| Multiple simile (*al-tashbīh al-mutaʿaddid*) – has mulitple simile features. | سليم كأسد في الشجاعة والجمال.<br>Salim is like a lion in courage and beauty. |
| Compound simile (*al-tashbīh al-murakkab*) – the simile feature is omitted and the likened is replaced with an image. | سليم كملك الحيوانات.<br>Salim is like the king of the animals. |
| Effective simile (*al-tashbīh al-balīgh*) – has the likened and the likened-to only. | سليم أسد.<br>Salim is a lion. |
| Reverse simile (*al-tashbīh al-maqlūb*) – the likened resembles the likened-to. | الأسد كسليم.<br>The lion is like Salim. |
| Implicit simile (*al-tashbīh al-ḍimniyy*) – only an attribute of the likened is mentioned. | لا يخشى الأسد أحداً.<br>A lion fears no-one. |
| Imaginary simile (*al-tashbīh al-wahamiyy*) – imaginary attributes are assigned to the likened. | سليم كأسد ذو رأسين.<br>Salim is like a lion with two heads. |

| | | |
|---|---|---|
| **Score** | This literary device is not present | **0** |
| | This literary device is present | **1** |
| | This literary device is present, with the highest rhetorical effect | **2** |

| **B-2: Metaphor**[37] | **الاستعارة** |
|---|---|

***Definition***

A metaphor is a figure of speech device which is derived from the "effective simile" which consists of the likened-to and the likened only. To make a metaphor, either the likened-to or the likened is omitted.

***Example***

سليم أسد.

Salim is a lion.

↓

دخل أسدٌ الغرفة.

A lion entered the room.

***Types of metaphor***

The types of metaphor include:

1) Explicit metaphor (*al-istiʿāra al-taṣrīḥiyya*) as in the example above where the likened-to is omitted.
2) Implicit metaphor (*al-istiʿāra al-makaniyyah*) where the likened is omited and something contextually symbolic from it is mentioned instead:

زأر الملك على الجموع.

The King roared at the crowds.

[37] H. Abdul-Raof, 2006, 218-225 & A. Abd Al-Ghanī, 2011, 67-86.

3) Enhanced metaphor (*al-istiʿāra al-murashshaḥa*) which is an implicit metaphor with additional description of the likened:

زأر الملك على الجموع وصعد إلى المنصة جلالة.

The King roared at the crowds as he majestically ascended to the podium.

4) Naked metaphor (*al-istiʿāra al-mujarrida*) in which an explicit metaphor is enhanced with lexical items appropriate to the likened:

دخل أسدٌ الغرفة وكان يرتدي بدلة راقية.

A lion entered the room and he was wearing an classy suit.

5) Absolute metaphor (*al-istiʿāra al-muṭlaqa*) which is an explicit metaphor which either does not include any lexical items relevant to the likened or the likened-to, or includes items which are relevant to both.
6) Proverbial metaphor (*al-istiʿāra al-tamthīliyya*) which draws attention to received wisdom in an allegorical way.

| | | |
|---|---|---|
| **Score** | This literary device is not present | **0** |
| | This literary device is present | **1** |
| | This literary device is present, with the highest rhetorical effect | **2** |

| **B-3: Allegory[38]** | **المجاز** |
|---|---|

***Definition & Examples***

Allegory is taking a word which has one meaning, and using it in a different but closely related way. There are two types of allegory:

1) Cognitive allegory (*al-majāz al-ʿaqlī*) where the subject of a verb is not the literal subject of that verb and the addressee is required to think about who the literal subject is. The allegorical subject and the real subject share a relationship which may take one of several forms:
   a) Causality (*al-ʿalāqa al-sababiyya*) relationship:

[38] H. Abdul-Raof, 2006, 226-232 & A. Abd Al-Ghanī, 2011, 121-150.

أعلنت الحكومة قواعد جديدة.

The government has announced new laws.

b) Time (*al-ʿalāqa al-zamāniyya*) relationship:

يمر الوقت بسرعة في العطلة.

Time passes fast in the holidays.

c) Place relationship (*al-ʿalāqa al-maḥaliyya):*

تلهم المدينةُ الإبداع.

The city inspires creativity.

2) Linguistic allegory (hypallage, *al-majāz al-mursal*) where this are contextual clues to the relationship between the allegorical and the real meanings. This relationship can take one of several forms:
   a) Synecdoche, where the whole is allegorically mentioned but only the part is intended literally, or vice-versa. There can be a whole-to-part relationship (*al-ʿalāqa al-kulliyya)* or a part-to-whole relationship (*al-ʿalāqa al-juzʾiyya*) between the allegorical and the literal:

رأيت لندن في عطلتي.

I saw London in my holiday.

تعال إلى بيتي لتناول الشاي الساعة 4 مساءً.

Come to my house for tea at 4pm.

   b) Place relationship (*al-ʿalāqa al-maḥaliyya)* where the place is mentioned but something related to the place is intended, or vice-versa.

يمكن أن تكون المحاكم في الهند فاسدة.

The courts in India can be corrupt.

زرتُ عائلة صديقي.

I visited my friend’s family.

c) Causality relationship (*al-ʿalāqa al-sababiyya*) where the cause is mentioned but effect is intended

فتح صديقي عيني على الفساد في المجتمع.

My friend opened my eyes to the corruption in society.

d) Effect relationship (*al-ʿalāqa al-musababiyya*) where the result is mentioned but the cause is intended

العمل الجيد يضع الطعام على الطاولة.

A good job puts food on the table.

e) Past relationship (*iʿtibar mā kāna*) where something from the past is mentioned in the present.

ترك طفلي المنزل للالتحاق بالجامعة.

My child left home to attend university.

f) Future relationship (*iʿtibar mā sayakūnu*) where something from the future is mentioned in the present

أزرعُ شجرة تفاح.

I am planting an apple tree.

g) Opposite relationship (*al-ʿalāqa ḍiddiyya*) where the opposite of what is said is intended

أتمنى أن تستمتع بحرياتك الجديدة بعد استقالتك.

I hope you enjoy your new freedoms, now that you have resigned.

h) Generalisation relationship (*al-ʿalāqa al-ʿumūm*) where a generalisation is made

الهنود فقراء.

Indians are poor.

i) Specific relationship (*al-ʿalāqa al-khuṣūṣiyya*) where a general thing is mentioned but a specific thing is intended.

مشى الأمريكيون على سطح القمر عام 1969.

| The Americans walked on the moon in 1969. | | |
|---|---|---|
| **Score** | This literary device is not present | **0** |
| | This literary device is present | **1** |
| | This literary device is present, with the highest rhetorical effect | **2** |

| **B-4: Metonymy / Implicit Reference[39]** | **الكناية** |
|---|---|

***Definition***

Metonymy is the use of a descriptive word or phrase which alludes to the intrinsic characteristics of a something, in place of the real name. Although the connection between the metonym and the real word is allegorical, it must be possible for the metonym to be literally true as well.

***Example***

سليم لسانه طويل.

Salim's tongue is long.

***Explanation***

Describing Salim's tongue as being long implies that he talks a lot, or that he has vulgar or deceitful speech. However, it could also mean that his tongue is literally quite long.

| | | |
|---|---|---|
| **Score** | This literary device is not present | **0** |
| | This literary device is present | **1** |
| | This literary device is present, with the highest rhetorical effect | **2** |

| **B-5: Hinting[40]** | **التعريض** |
|---|---|

***Definition***

[39] H. Abdul-Raof, 2006, 233-238 & A. Abd Al-Ghanī, 2011, 93-106.
[40] A. Abd Al-Ghanī, 2011, 107-112

| |
|---|
| This is indirect suggestion of the intended meaning through the use of an aphorism, a proverb, a riddle or an innuendo, rather than directly stating the intended message. |

| | | |
|---|---|---|
| **Score** | This literary device is not present | **0** |
| | This literary device is present | **1** |
| | This literary device is present, with the highest rhetorical effect | **2** |

| **B-6: Pun / Paronomasia / Double-Entendre**[41] | التورية |
|---|---|

***Definition***

This is the use of an expression that has more than one meaning. The speaker says one thing, but there is more than one interpretation of the meaning. This causes a joking, ironic, sarcastic or mocking effect due to the deliberate confusion caused, between the superficial meaning, and the construed meaning. The speaker technically speaking does not lie, as both interpretations are correct.

***See also***

- **CC-3 Complete *Jinās***, where the same word is used twice, but with a different meaning each time.

| | | |
|---|---|---|
| **Score** | This literary device is not present | **0** |
| | This literary device is present | **1** |
| | This literary device is present, with the highest rhetorical effect | **2** |

[41] A. Abd Al-Ghanī, 2011, 275-290.

### 3.3) Domain C: Embellishments

*3.3.1) PART A – Word Choice*

| **CA-1: Meaningful Proper Nouns**[42] | | **التوجيه** |
|---|---|---|
| ***Definition***<br>Use of names of people or places that have meanings / can have multiple meanings / interpretations. | | |
| **Score** | This literary device is not present | **0** |
| | This literary device is present | **1** |
| | This literary device is present, with the highest rhetorical effect | **2** |

| **CA-2: Oxymoron**[43] | | **الارداف الخلفي / اجتماع لفظتين متناقضتين** |
|---|---|---|
| ***Definition***<br>The co-location of two antonyms.<br>***Example***<br>نعيش في قرية عالمية.<br>We live in a global village. | | |
| **Score** | This literary device is not present | **0** |
| | This literary device is present | **1** |
| | This literary device is present, with the highest rhetorical effect | **2** |

42 H. Nasif et al, 2012, 106-107.
43 H. Abdul-Raof, 2006, 254.

| CA-3: Amphibology[44] | الإبهام |
|---|---|

**_Definition_**

Use of a word or phrase ambiguously such that can have 2 opposite meanings can be understood.

| | | |
|---|---|---|
| Score | This literary device is not present | 0 |
| | This literary device is present | 1 |
| | This literary device is present, with the highest rhetorical effect | 2 |

| CA-4: Onomatopoeia[45] | المحاكاة الصوتية |
|---|---|

**_Definition_**

The representation of sounds where the pronunciation of the word mimics the sound itself.

| | | |
|---|---|---|
| Score | This literary device is not present | 0 |
| | This literary device is present | 1 |
| | This literary device is present, with the highest rhetorical effect | 2 |

| CA-5: Litotes[46] | الإثبات بالنفي |
|---|---|

**_Definition_**

To allude to a word by negating its antonym.

**_Example_**

ليست لندن مدينة صغيرة!

London is not a small city!

---

[44] H. Nasif et al, 2012, 106.
[45] H. Abdul-Raof, 2006, 267.
[46] H. Abdul-Raof, 2006, 252.

| Score | | |
|---|---|---|
| | This literary device is not present | **0** |
| | This literary device is present | **1** |
| | This literary device is present, with the highest rhetorical effect | **2** |

| **CA-6: Alliteration[47]** | **المجانسة الاستهلاكية** |
|---|---|

***Definition***

The appearance of a succession of words where the initial letter of the word is same.

***Example***

وَمَكَرُواْ مَكْرًا وَمَكَرْنَا مَكْرًا وَهُمْ لَا يَشْعُرُونَ

Quran 27 :50

They planned a plan, and We planned a plan

| Score | | |
|---|---|---|
| | This literary device is not present | **0** |
| | This literary device is present | **1** |
| | This literary device is present, with the highest rhetorical effect | **2** |

| **CA-7: Palindrome[48]** | **القلب / ما لا يستحيل بالانعكاس** |
|---|---|

***Definition***

The communicator uses a lexical item that can be read both forwards and backwards.

| Score | | |
|---|---|---|
| | This literary device is not present | **0** |
| | This literary device is present | **1** |
| | This literary device is present, with the highest rhetorical effect | **2** |

47 H. Abdul-Raof, 2006, 259-260.

48 H. Nasif et al, 2012, 120 & H. Muhammad, 2016, 61-62.

| CA-8: Equivocation[49] | | | المواربة |
|---|---|---|---|
| ***Definition***<br><br>The communicator uses a word which can change its meaning through a small change in the word itself.<br><br>***See also***<br><br>➢ **CC-3: Incomplete *Jinās*** | | | |
| **Score** | This literary device is not present | | **0** |
| | This literary device is present | | **1** |
| | This literary device is present, with the highest rhetorical effect | | **2** |

| CA-9: Adornment[50] | | | التدبيج |
|---|---|---|---|
| ***Definition***<br><br>The use of contrasting colours such as red and green or black and white. | | | |
| **Score** | This literary device is not present | | **0** |
| | This literary device is present | | **1** |
| | This literary device is present, with the highest rhetorical effect | | **2** |

| CA-10: Metabole[51] | التكرار بعبارات مختلفة |
|---|---|
| ***Definition***<br><br>This is the modification of a lexical item by two or more lexical items.<br><br>***Example***<br><br>سليم مؤدب ولطيف وصبور. | |

[49] H. Nasif et al, 2012, 121.
[50] H. Nasif et al, 2012, 107.
[51] H. Abdul-Raof, 2006, 267.

| |
|---|
| Salim is polite, kind and patient.<br><br>***Explanation***<br><br>The noun “Salim” is modified by the descriptors “polite,” “kind” and “patient”. |

| | | |
|---|---|---|
| **Score** | This literary device is not present | **0** |
| | This literary device is present | **1** |
| | This literary device is present, with the highest rhetorical effect | **2** |

| **CA-11: Zeugma[52]** | **العبارة الجامعة** |
|---|---|

***Definition***

A word or expression is used to modify two other words, each in different ways.

***Example***

فتحت له بابها وقلبها.

She opened her door and her heart to him.

***Explanation***

The word “opened” has a literal meaning with opening the door, and an allegorical meaning with opening her heart.

| | | |
|---|---|---|
| **Score** | This literary device is not present | **0** |
| | This literary device is present | **1** |
| | This literary device is present, with the highest rhetorical effect | **2** |

| **CA-12: Al-Istikhdām[53]** | **الاستخدام** |
|---|---|

***Definition***

[52] H. Abdul-Raof, 2006, 269-270.

[53] H. Nasif et al, 2012, 108-109.

The communicator mentions something and then uses a personal pronoun or demonstrative pronoun to apparently refer back to that thing. However, something related but different is intended by the pronoun.

***Example***

فَمَن شَهِدَ مِنكُمُ ٱلشَّهْرَ فَلْيَصُمْهُ

Quran 2: 185

So whoever is present this month, let them fast it.

***Explanation***

The word *shahar* (month) refers to the new crescent moon signifying the start of the month of Ramadan. Whereas the *hu* (it) refers to the fasting in the days of Ramadan, and not the crescent moon itself.

| | | |
|---|---|---|
| **Score** | This literary device is not present | **0** |
| | This literary device is present | **1** |
| | This literary device is present, with the highest rhetorical effect | **2** |

| **CA-13: Epizeuxis**[54] | **التكرار التوكيدي / التوكيد اللفظي** |
|---|---|

***Definition***

The repetition of the same word or expressions for affirmation.

| | | |
|---|---|---|
| **Score** | This literary device is not present | **0** |
| | This literary device is present | **1** |
| | This literary device is present, with the highest rhetorical effect | **2** |

[54] H. Abdul-Raof, 2006, 250-251.

| CA-14: Epistrophe[55] | | | تكرار النهاية |
|---|---|---|---|
| ***Definition***<br>Repetition of the same word or expressions at the end of the sentence for affirmation. | | | |
| **Score** | This literary device is not present | **0** | |
| | This literary device is present | **1** | |
| | This literary device is present, with the highest rhetorical effect | **2** | |

*3.3.2) PART B – Addressing Groups*

| CB-1: Congeries[56] | | | مراعاة النظير |
|---|---|---|---|
| ***Definition***<br>This is the amplification of rhetorical effect by collecting together of discrete lexical items, which are related to the thing being referred to.<br>***Example***<br>يقع هذا البيت في حي رائع، في حالة جيدة لعمره، قريب من مدرسة متميزة، والسعر معقول. هو مناسب تماماً لنا!<br>This house is located in a great neighbourhood, in good condition for its age, close to an excellent school, and the price is reasonable. It is perfect for us! | | | |
| **Score** | This literary device is not present | **0** | |
| | This literary device is present | **1** | |
| | This literary device is present, with the highest rhetorical effect | **2** | |

[55] H. Abdul-Raof, 2006, 249-250.
[56] H. Nasif et al, 2012, 108.

| CB-2: Collectiveness[57] | الجمع |
|---|---|

***Definition***

This is to combine several things together and give a collective verdict or judgement on them together.

***Example***

دفاترك وأقلامك وكل جهدك ستكون ضرورية عند بدء العام الدراسي الجديد.

Your notebooks, pens, and all your efforts will be essential when the new school year begins.

| | | |
|---|---|---|
| **Score** | This literary device is not present | **0** |
| | This literary device is present | **1** |
| | This literary device is present, with the highest rhetorical effect | **2** |

| CB-3: Al-Taqsīm[58] | التقسيم |
|---|---|

***Definition***

This is when all the constituent members of a group are mentioned, or when each constituent member of a group is mentioned along with something specific about each one.

***Example***

يجب عليك شراء الدفاتر الجديدة، وترتيب مكتبك، والعثور على كل أقلامك قبل بدء العام الدراسي الجديد.

You have to buy new notebooks, tidy your desk, and find all your pens before the new school year begins.

| | | |
|---|---|---|
| **Score** | This literary device is not present | **0** |
| | This literary device is present | **1** |

[57] H. Nasif et al, 2012, 110.
[58] H. Nasif et al, 2012, 111.

| | | |
|---|---|---|
| | This literary device is present, with the highest rhetorical effect | **2** |

| **CB-4: Differentiation Between Similar Items**[59] | **التفريق** |
|---|---|

***Definition***

This is to differentiate between two things that mistakenly could be considered to be same.

***Example***

سليم وكريم طالبان مختلفان تمامًا: سالم موهوب بالفطرة، ولكن على كريم أن يدرس بجد إذا أراد نتائج جيدة.

Salim and Karim are very different students: Salim is naturally gifted but Karim has to study hard if he wants good results.

| | | |
|---|---|---|
| **Score** | This literary device is not present | **0** |
| | This literary device is present | **1** |
| | This literary device is present, with the highest rhetorical effect | **2** |

| **CB-5: Epanodos**[60] | **الطي والنشر / اللف والنشر** |
|---|---|

***Definition***

The communicator mentions two or more items in succession. Then, further information is given about these two items. The addressee understands which piece of information corresponds with which item, without the communicator having to specify this explicitly.

***Example***

الشمس والقمر مختلفان تمامًا: أحدهما يخلق الضوء، والآخر يعكسه فقط.

The sun and moon are very different: one creates light, the other merely reflects it.

[59] H. Nasif et al, 2012, 110.
[60] H. Abdul-Raof, 2006, 248-250 & H. Nasif et al, 2012, 111-112.

| | | |
|---|---|---|
| **Score** | This literary device is not present | **0** |
| | This literary device is present | **1** |
| | This literary device is present, with the highest rhetorical effect | **2** |

### *3.3.3) PART C – Sentence Construction*

| **CC-1: Antithesis / Antonymy[61]** | **الطباق / المقابلة** |
|---|---|

***Definition***

This is where two words with opposite meanings (antonyms) appear together, such as “big” and “small.” The two words may be allegorical or non-allegorical. The antithesis construct may be either non-negated antithesis using direct opposites such as “true” and “false,” or negated antithesis using negation of the same word (such as “true” and “not true”). The antithesis may occur between two nouns, two verbs, two particles, or any combination of these. Multiple antithesis may occur where two or more antonyms or antonymic phrases appear together.

***Example***

فَلْيَضْحَكُوا۟ قَلِيلًا وَلْيَبْكُوا۟ كَثِيرًا جَزَآءًۢ بِمَا كَانُوا۟ يَكْسِبُونَ

Quran 9:82

So let them laugh a little—they will weep much as a reward for what they have committed.

***Explanation***

There is antithesis between *yaḍḥakū* (laugh) and *yabkū* (cry), and between *qalīl* (a little) and *kathīr* (a lot).

***Scoring criterion***

[61] H. Abdul-Raof, 2006, 245-246 & 252-254.
A. Abd Al-Ghanī, 2011, P171-178, & 183-188.

| | | |
|---|---|---|
| One or two marks should be award for each pair of antonyms or antonymic phrases. For example, if there are three pairs of antonymic phrases, a mark of 3 or 6 should be awarded. | | |
| **Score** | This literary device is not present | **0** |
| | This literary device is present | **1** |
| | This literary device is present, with the highest rhetorical effect | **2** |

| **CC-2: Chiasmus / Antimetabole[62]** | **المقابلة العكسية** |
|---|---|

***Definition***

This is a two-part sentence or phrase, where the word order in the first part is reversed in the second part.

***Example***

يُخْرِجُ ٱلْحَىَّ مِنَ ٱلْمَيِّتِ وَيُخْرِجُ ٱلْمَيِّتَ مِنَ ٱلْحَىِّ

He brings the living out of the dead and brings the dead out of the living

Quran 30:19

| | | |
|---|---|---|
| **Score** | This literary device is not present | **0** |
| | This literary device is present | **1** |
| | This literary device is present, with the highest rhetorical effect | **2** |

| **CC-3: Al-Jinās[63]** | **الجناس / التجنيس** |
|---|---|

***Definition***

The occurrence of two words which sound similar but have different meanings. There are several types of *jinās*:

[62] H. Abdul-Raof, 2006, 247-248.
[63] H. Abdul-Raof, 2006, 261-267 & A. Abd Al-Ghanī, 2011, 223-234.

1) Complete *jinās* (*al-jinās al-tāmm*) – the two words are identical but they have different meanings. This is also known as a pun.
2) Incomplete *jinās* (*al-jinās ghayr al-tāmm*) – the two words differ in one aspect such as:
    a) Reverse *jinās* (*jinās al-qalb*): the letter order is changed to make an anagram or a palindrome.
    b) Morphological *jinās* (*jinās al-ishtiqāq*): the two words are derived from the same root but have different morphological forms.
    c) Fabricated *jinās (al-jinās al-mulaffaq)*: two unrelated words that sound the same appear together.
    d) Resemblance *jinās (al-jinās al-muḍariʿ)*: the two words differ by one letter, but the different letters have the same point of articulation.
    e) Non-resemblance *jinās (al-jinās al-lāḥiq)*: the two words differ by one letter, and the different letters have different points of articulation.
    f) Distorted *jinās (al-jinās al-muḥarraf)*: the two words are the same but have a difference in a vowel.

| | | |
|---|---|---|
| **Score** | This literary device is not present | **0** |
| | This literary device is present | **1** |
| | This literary device is present, with the highest rhetorical effect | **2** |

| **CC-4: Tail-Head**[64] | رد العجز على الصدر / التصدير |
|---|---|

***Definition***

The first word of the sentence and the last word of the sentence are the same or morphologically related.

***Example***

[64] H. Abdul-Raof, 2006, 268-269 & H. Nasif et al, 2012, 117.

| |
|---|
| افعلْ ما تشاء؛ ليست مسؤوليتي أن أقول لك ما عليك أن تفعل.<br>Do what you like; it is not my responsibility to tell you what you must do. |

| | | |
|---|---|---|
| **Score** | This literary device is not present | **0** |
| | This literary device is present | **1** |
| | This literary device is present, with the highest rhetorical effect | **2** |

| **CC-5: Head-Tail[65]** | **رد الصدر على العجز** |
|---|---|

***Definition***

The last word of the first sentence and the first word of the second sentence are the same or morphologically related.

***Example***

لندن مدينة الفرص.

الفرص المفتوحة للاستغلال.

والاستغلال وباء.

والوباء لا يميز بين الغني والفقير.

London is a city of opportunity.

Opportunities open for exploitation.

Exploitation is an epidemic.

And the epidemic does not distinguish between rich and poor.

| | | |
|---|---|---|
| **Score** | This literary device is not present | **0** |
| | This literary device is present | **1** |
| | This literary device is present, with the highest rhetorical effect | **2** |

65 H. Abdul-Raof, 2006, 261.

| CC-6: Similarities of the Start & Finish[66] | تشابه الأطراف |
|---|---|

***Definition***

The end of a sentence has a similar meaning to the beginning of the following sentence, establishing a semantic link which bridges the two sentences.

***Example***

يريد الجميع شراء منزل.

شراء البيت أو الشقة هو أكبر عملية شراء في حياتك.

These days, everyone wants to buy a home.

A house or flat is biggest purchase in your life.

| | | |
|---|---|---|
| **Score** | This literary device is not present | **0** |
| | This literary device is present | **1** |
| | This literary device is present, with the highest rhetorical effect | **2** |

| CC-7: Parallelism[67] | الموازنة / مقابلة اللفظ باللفظ |
|---|---|

***Definition***

The communicator repeats the lexical structure of a sentence or phrase over two or more sentences to create assonance and rhyme.

| | | |
|---|---|---|
| **Score** | This literary device is not present | **0** |
| | This literary device is present | **1** |
| | This literary device is present, with the highest rhetorical effect | **2** |

66 H. Nasif et al, 2012, 116.
67 H. Abdul-Raof, 2006, 268-269.

*3.3.4) PART D – Musicality*

| **CD-1: Assonance[68]** | | | السجع |
|---|---|---|---|
| ***Definition***<br>The communicator creates rhyme by the agreement of vowel endings of the final words of two sentences. The assonance may be of three grades:<br>• Low: The final word of each sentence a has different morphology.<br>• Intermediate: The final word of each sentence has the same morphology<br>• High: The words of both sentences have the same morphology | | | |
| **Score** | This literary device is not present | **0** | |
| | This literary device is present | **1** | |
| | This literary device is present, with the highest rhetorical effect | **2** | |

| **CD-2: Homeoptoton[69]** | | | الترصيع / المرادف |
|---|---|---|---|
| ***Definition***<br>The final part of the first hemistich of a line of poetry agrees with final part of the second hemistich in metre, vowelling and rhyme. | | | |
| **Score** | This literary device is not present | **0** | |
| | This literary device is present | **1** | |
| | This literary device is present, with the highest rhetorical effect | **2** | |

[68] H. Abdul-Raof, 2006, 260-261.
A. Abd Al-Ghanī, 2011, 193-199.
H. Nasif et al, 2012, 119-120.
[69] A. Abd Al-Ghanī, 2011, 241-244.

| CD-3: Concordance of the Pronunciation and the Meaning[70] | ائتلاف اللفظ والمعنى |
|---|---|

***Definition***

There is agreement between the sounds of the words and their meanings. For example, words with soft sounds are used to convey a soft or delicate tone. Conversely, words with harsh sounds are used to convey a harsh tone.

| | | |
|---|---|---|
| **Score** | This literary device is not present | **0** |
| | This literary device is present | **1** |
| | This literary device is present, with the highest rhetorical effect | **2** |

| CD-4: Concordance of Pronunciations[71] | ائتلاف اللفظ مع اللفظ |
|---|---|

***Definition***

The communicator brings together strange, unusual and unexpected words for rhetorical effect.

***Examples***

قَالُواْ تَٱللَّهِ تَفْتَؤُاْ تَذْكُرُ يُوسُفَ حَتَّىٰ تَكُونَ حَرَضًا أَوْ تَكُونَ مِنَ ٱلْهَـٰلِكِينَ

Quran 12: 85

They said, “By Allah! You will not cease to remember Joseph until you lose your health or ˹even˺ your life.”

***Explanation***

Both (i) the use of *ta-* as an oath device prefixed to “Allah,” and (ii) the use of the word *taftaʾu* (to continue), are rarely used in Arabic, but are co-located in this text.

| | | |
|---|---|---|
| **Score** | This literary device is not present | **0** |
| | This literary device is present | **1** |

[70] H. Nasif et al, 2012, 115-6.
[71] H. Nasif et al, 2012, 121.

| | | |
|---|---|---|
| | This literary device is present, with the highest rhetorical effect | **2** |

| **CD-5: Al-Tashrīʿ [72]** | **التشريع** |
|---|---|

***Definition***

The rhyme and or poetic metre of a poem is such that even if some words are omitted, the rhyme or metre are maintained.

| | | |
|---|---|---|
| **Score** | This literary device is not present | **0** |
| | This literary device is present | **1** |
| | This literary device is present, with the highest rhetorical effect | **2** |

| **CD-6: Proportioning [73]** | **الازدواج** |
|---|---|

***Definition***

The communicator divides a sentence or paragraph into sentences with equal length and poetic metre.

| | | |
|---|---|---|
| **Score** | This literary device is not present | **0** |
| | This literary device is present | **1** |
| | This literary device is present, with the highest rhetorical effect | **2** |

| **CD-7: Excellence of Division[74]** | **حسن التقسيم** |
|---|---|

***Definition***

The communicator divides each line of a poem into two equal hemi-stiches.

| | | |
|---|---|---|
| **Score** | This literary device is not present | **0** |
| | This literary device is present | **1** |

[72] H. Nasif et al, 2012, 120-121.
[73] A. Abd Al-Ghanī, 2011, 213-217.
[74] A. Abd Al-Ghanī, 2011, 249-251.

| | | |
|---|---|---|
| | This literary device is present, with the highest rhetorical effect | **2** |

### *3.3.5) PART E – Strengthening the Argument*

| **CE-1: Integration of Imagery[75]** | **الإدماج** |
|---|---|

***Definition***

The communicator uses imagery to imply the existence of something else, rather than explicitly mentioning it.

***Example***

كنت أنظر إلى الساعة طوال الليل، في انتظار الفجر المتأخر.

I was looking at the clock all night, waiting for the dawn.

***Explanation***

The imagery implies that the narrator was suffering from insomnia.

| | | |
|---|---|---|
| **Score** | This literary device is not present | **0** |
| | This literary device is present | **1** |
| | This literary device is present, with the highest rhetorical effect | **2** |

| **CE-2: Stacked-up Descriptions [76]** | **الاستتبع** |
|---|---|

***Definition***

Rather than only one statement of praise or criticism, the communicator links together two or more related statements of praise or criticism. This creates a more intense rhetorical effect.

***Example***

لقد حصلت على درجات ممتازة في امتحاناتي النهائية،

لقد ضمنت مستقبلك المشرق،

[75] H. Nasif et al, 2012, 107.

[76] H. Nasif et al, 2012, 107-108.

| |
|---|
| وأنت دحضت كل المشككين.<br><br>You got excellent grades in your final exams,<br>You secured your bright future,<br>And you disproved all the sceptics. |

| | | |
|---|---|---|
| **Score** | This literary device is not present | **0** |
| | This literary device is present | **1** |
| | This literary device is present, with the highest rhetorical effect | **2** |

| **CE-3: Incorporation of Proverbs**[77] | **إرسال المثل / الكلام الجامع** |
|---|---|

***Definition***

The communicator incorporates a proverb, parable, or well-known statement into the text to strengthen the argument.

***Example***

إذا جاء موسى وألقى العصا... فقد بطل السحر والساحر.

When Moses came and he threw down the stick… the magic and the magicians were defeated.

***Explanation***

This is a well-known proverb which refers to the story of Moses in Abrahamic texts, when he threw down his stick which became a snake that ate what Pharoah's magicians were depicting as snakes. It alludes to the fact that truth will conquer falsehood. Incorporation of this proverb into a piece will strengthen its rhetorical strength.

| | | |
|---|---|---|
| **Score** | This literary device is not present | **0** |
| | This literary device is present | **1** |
| | This literary device is present, with the highest rhetorical effect | **2** |

[77] H. Nasif et al, 2012, 112.

| CE-4: Abstraction[78] | التجريد |
|---|---|

***Definition***

The communicator is mentioning a proposition, and side-tracks to mention something else that exaggerates or epitomises the quality the original proposition has.

***Example***

هذه الجامعة من أفضل الجامعات في البلاد بالنسبة للطالب. فاز نائب المستشار مؤخرًا بجائزة نوبل.

For a student, this is one of the best universities in the country. The Vice-Chancellor recently won a Nobel prize.

***Explanation***

The Vice-Chancellor winning a Nobel Prize does not directly affect the quality of teaching but is a reflection of the excellence of the institution overall.

| | | |
|---|---|---|
| **Score** | This literary device is not present | **0** |
| | This literary device is present | **1** |
| | This literary device is present, with the highest rhetorical effect | **2** |

| CE-5: Quotation[79] | الاقتباس |
|---|---|

***Definition***

The communicator takes a well-known text such as a proverb, poem, story, prophetic sayings, or Quranic verse and incorporates it into a new text.

| | | |
|---|---|---|
| **Score** | This literary device is not present | **0** |
| | This literary device is present | **1** |
| | This literary device is present, with the highest rhetorical effect | **2** |

[78] H. Nasif et al, 2012, 115.
[79] H. Abdul-Raof, 2006,256 & A. Abd Al-Ghanī, 2011, 307-314.

| CE-6: Hinting at the Source[80] | التلميح |
|---|---|

***Definition***

When the communicator takes a well-known text such as a proverb, poem, story, prophetic sayings, or Quranic verse and incorporates it into a new text, the communicator provides a clue about the source of the quotation.

| | | |
|---|---|---|
| **Score** | This literary device is not present | **0** |
| | This literary device is present | **1** |
| | This literary device is present, with the highest rhetorical effect | **2** |

| CE-7: Euphemism[81] | التهوين |
|---|---|

***Definition***

The communicator uses an implicit reference rather than explicitly referring to something unpleasant or embarrassing.

***Example***

ذهب كريم إلى الجانب المظلم.

Karim has gone to the dark side.

| | | |
|---|---|---|
| **Score** | This literary device is not present | **0** |
| | This literary device is present | **1** |
| | This literary device is present, with the highest rhetorical effect | **2** |

| CE-8: Rhetorical Shift[82] | الالتفات |
|---|---|

***Definition***

The communicator changes between:

[80] H. Nasif et al, 2012, 127-128.
[81] H. Abdul-Raof, 2006, 251.
[82] H. Abdul-Raof, 2006, 257-258 & A. Abd Al-Ghanī, 2011, 255-260.

- Persons: first person, second person, third person
- Tenses: past tense, present tense, future, imperative
- Morphology: verb, active participle, passive participle
- Sentence type: verbal sentence, nominal sentence

This shift keeps the addressee engaged and provides stylistic diversity.

***Example***

When Salim has lost his key:

سليم: أين مفتاحي؟ كان في جيبي. أين وضعت المفتاح يا سليم؟

كيف يفقد الرجل مفتاح بيته؟ تعال يا سالم ... جد المفتاح بسرعة!

Salim: Where is my key? It was in my pocket. Where did you put the key, Salim? How does a man lose his house key? Come on Salim... find the key quickly!

***Explanation***

Salim refers to himself in the first person ("my key"), then the second person ("you"), and then the third person ("a man"). There are changes in verb from past tense ("was"), to present tense ("does") to imperative ("find").

| | | |
|---|---|---|
| **Score** | This literary device is not present | **0** |
| | This literary device is present | **1** |
| | This literary device is present, with the highest rhetorical effect | **2** |

| **CE-9: Epitrope[83]** | **التسليم الخطابي** |
|---|---|

***Definition***

The communicator apparently accepts the opponent's argument, but then provides the antithesis to it.

***Example***

[83] H. Abdul-Raof, 2006, 250.

| |
|---|
| كريم: هل أنفقت كل أموالنا؟<br>سليم: نعم أنفقت كل المال. لكننا كنا جائعين أثناء الرحلة ولم يكن لدي خيار آخر.<br><br>Karim: Did you spend all our money?<br>Salim: Yes, I spent all the money. But we were hungry during the journey, and I had no other choice. |

| | | |
|---|---|---|
| **Score** | This literary device is not present | **0** |
| | This literary device is present | **1** |
| | This literary device is present, with the highest rhetorical effect | **2** |

| **CE-10: Evasive Response[84]** | **أسلوب الحكيم** |
|---|---|
| ***Definition***<br>The communicator responds to a question with an evasive or ambiguous response. This evokes an effect in the addressee.<br>***Example***<br>كريم: كم عمرك؟<br>سليم: لماذا تريد معرفة هذه المعلومة الشخصية؟<br><br>Karim: How old are you?<br>Salim: Why do you want this personal information? | |

| | | |
|---|---|---|
| **Score** | This literary device is not present | **0** |
| | This literary device is present | **1** |
| | This literary device is present, with the highest rhetorical effect | **2** |

[84] H. Abdul-Raof, 2006, 251.

| CE-11: Feigned Ignorance[85] | تجاهل العارف |
|---|---|

***Definition***

The communicator knows something but pretends not to know. This literary device may be found when the communicator wishes to express amazement, praise, dispraise, disapproval, or to reprimand, or create a sense of familiarity.

***Example***

The communicator wishes to impart some advice about starting a company:

أتمنى أنني كنت أعرف هذا الأمر عندما أسست هذه الشركة قبل 10 سنوات.

"This is something I wish I knew when I started this company 10 years ago."

***Explanation***

It is possible that the information was actually known when the company was started, but feigning ignorance has been used to setup a rhetorical image of ignorance which mirrors the addressee's condition.

| | | |
|---|---|---|
| **Score** | This literary device is not present | **0** |
| | This literary device is present | **1** |
| | This literary device is present, with the highest rhetorical effect | **2** |

| CE-12: Observation[86] | الإرصاد / التسهيم |
|---|---|

***Definition***

The communicator provides the addressee with hints at the start of the text such that the addressee can predict what the communicator will say.

***Example***

[85] H. Muhammad, 2016, 48.

[86] H. Abdul-Raof, 2006,253-254.

| |
|---|
| كنت سعيدًا حقًا بنتائج هذا الامتحان. مبروك عليكم!<br>لقد حصل كل منكم أكثر من ٩٥% درجة.<br>"I was really very happy with the results of this exam. Well done!<br>All of you have scored more than 95%."<br><br>***Explanation***<br><br>The teacher hints at the good marks prior to stating that everyone got more than 95%. |

| | | |
|---|---|---|
| **Score** | This literary device is not present | **0** |
| | This literary device is present | **1** |
| | This literary device is present, with the highest rhetorical effect | **2** |

| **CE-13: Apostrophe[87]** | **مخاطبة غير العاقل** |
|---|---|

***Definition***

This is a form of personification where the communicator directly addresses a non-human object as if it was human.

| | | |
|---|---|---|
| **Score** | Not present | **0** |
| | Present in a basic way | **1** |
| | Present in a way that significantly adds value | **2** |

| **CE-14: Personification[88]** | **تشخيص، تجسيد** |
|---|---|

***Definition***

The communicator refers to non-human objects as if they were human.

| | | |
|---|---|---|
| **Score** | Not present | **0** |
| | Present in a basic way | **1** |

87 H. Abdul-Raof, 2006,246-247.
88 H. Abdul-Raof, 2006, 255-256.

| | | |
|---|---|---|
| | Present in a way that significantly adds value | **2** |

| **CE-15 Hyperbole[89]** | **المبالغة** |
|---|---|

***Definition & examples***

The communicator exaggerates in describing a proposition. There are three grades:

1) Lowest: The proposition is conceptually possible and may normally occur.

هذه السيارة من أسرع السيارات التي يمكنك شراؤها بهذا السعر.

"This car is one of the fastest cars you can buy for this price."

2) Intermediate: The proposition is conceptually possible but does not usually occur.

هذه هي أسرع سيارة في العالم.

"This is the fastest car in the world."

3) Highest: The proposition is conceptually impossible and it does not happen.

يمكنك السفر بسرعة مليون ميل في الساعة في هذه السيارة.

"You'll be travelling at a million miles an hour in this car."

| | | |
|---|---|---|
| **Score** | This literary device is not present | **0** |
| | This literary device is present | **1** |
| | This literary device is present, with the highest rhetorical effect | **2** |

[89] H. Abdul-Raof, 2006, 252 & H. Nasif et al, 2012, 112-113.

| CE-16: Beauty of Rationale / Fantastic Aetiology / Conceit[90] | حسن التعليل |
|---|---|

***Definition***

The communicator denies the obvious cause of something happening, and instead proposes a different but fanciful cause. This has the effect of attributing the occurrence to something which benefits the communicator's objectives and agenda.

***Example***

لقد حققت النجاح بسبب دعاء والدتك الدائم لك فقط.

You achieved success only because of your mother's constant supplication for you.

| Score | | |
|---|---|---|
| | This literary device is not present | **0** |
| | This literary device is present | **1** |
| | This literary device is present, with the highest rhetorical effect | **2** |

| CE-17: Asteism / Affirmed Praise[91] | تأكيد المدح بما يُشبه الذم |
|---|---|

***Definition***

The communicator praises, then indicates that there is an exception to the praise, and then affirms the original praise by praising again. This is the opposite of Affirmed Dispraise.

***Example***

سليم طالب ممتاز. المشكلة الوحيدة هي أنه يستعد للدرس جيدًا لدرجة أنه يعرف بالفعل جميع الإجابات!

Salim is an excellent student. The only problem is that he prepares for the lesson so well that he already knows all the answers!

90 H. Abdul-Raof, 2006, 248 & H. Nasif et al, 2012, 115.
91 H. Abdul-Raof, 2006, 247 & A. Abd Al-Ghanī, 2011, 265-270.

| | | |
|---|---|---|
| | | |
| **Score** | This literary device is not present | **0** |
| | This literary device is present | **1** |
| | This literary device is present, with the highest rhetorical effect | **2** |

| **CE-18: Affirmed Dispraise**[92] | **تأكيد الذم بما يُشبه المدح** |
|---|---|

***Definition***

The communicator criticises, then indicates that there is an exception to the criticism, and then affirms the original criticism by criticising again. This is the opposite of Asteism / Affirmed Praise.

***Example***

كريم طالب سيئ. الشيء الوحيد الجيد فيه أنه يعرف مدى سوء حالته، ولا ينوي دراسة هذا الموضوع بعد هذا العام الدراسي.

"Karim is a bad student. The only thing good about him is that he knows how bad he is, and he does not intend to study this subject after this academic year."

| | | |
|---|---|---|
| **Score** | This literary device is not present | **0** |
| | This literary device is present | **1** |
| | This literary device is present, with the highest rhetorical effect | **2** |

| **CE-19: Al-Mughāyra**[93] | **المغايرة** |
|---|---|

***Definition***

*Al-Mughāyra* is when a praise is followed by a criticism, or vice-versa.

| | | |
|---|---|---|
| **Score** | This literary device is not present | **0** |

[92] H. Abdul-Raof, 2006, 245 & A. Abd Al-Ghanī, 2011, 265-270.

[93] H. Nasif et al, 2012, 113.

| | | |
|---|---|---|
| | This literary device is present | **1** |
| | This literary device is present, with the highest rhetorical effect | **2** |

| **CE-20: Tapinosis[94]** | التحقير |
|---|---|

***Definition***

The communicator depreciates someone or something by referring to it using terms which suggest it is less important or impressive than it is.

***Example***

أعيش في كوخ في الريف.

I live in a hut in the countryside.

| | | |
|---|---|---|
| **Score** | This literary device is not present | **0** |
| | This literary device is present | **1** |
| | This literary device is present, with the highest rhetorical effect | **2** |

| **CE-21: Sarcasm[95]** | الاستهزاء |
|---|---|

***Definition***

The communicator indirectly criticises, rebukes, or refutes by mentioning something positive which the addressee can understand or infer as actually being negative.

| | | |
|---|---|---|
| **Score** | This literary device is not present | **0** |
| | This literary device is present | **1** |
| | This literary device is present, with the highest rhetorical effect | **2** |

[94] H. Abdul-Raof, 2006,258-259.
[95] H. Abdul-Raof, 2006,256.

| CE-22: Scholastic Approach / Dialectical Mannerism[96] | المذهب الكلامي |
|---|---|

**_Definition_**

The communicator uses reasoning, rational ideas, logic, or evidence to convince the addressee of the communicator's proposition or to refute conflicting views.

| Score | | |
|---|---|---|
| | This literary device is not present | 0 |
| | This literary device is present | 1 |
| | This literary device is present, with the highest rhetorical effect | 2 |

*3.3.6) PART F – Paragraph Construction*

| CF-1: Multi-Genre[97] | الافتنان |
|---|---|

**_Definition_**

The text combines two or more different literary arts (such as eulogy, praise, satire, boasting, congratulation, condolence, reprimand) in one piece.

| Score | | |
|---|---|---|
| | This literary device is not present | 0 |
| | This literary device is present | 1 |
| | This literary device is present, with the highest rhetorical effect | 2 |

| CF-2: Pleasantness of the Opening[98] | حسن الابتداء |
|---|---|

**_Definition_**

The communicator employs a pleasant and agreeable opening to the text which pleases and addressee.

---

[96] H. Abdul-Raof, 2006, 257.
[97] H. Nasif et al, 2012, 109-110.
[98] H. Nasif et al, 2012, 128.

| | | |
|---|---|---|
| | | |
| **Score** | This literary device is not present | **0** |
| | This literary device is present | **1** |
| | This literary device is present, with the highest rhetorical effect | **2** |

| **CF-3: Exordium / Finesse of Initiation**[99] | **براعة الاستهلال** |
|---|---|

***Definition***

The communicator starts the text in an excellent manner which hints at the main objective of the text, attracts the attention of the addressee and makes the addressee eager to hear the main proposition of the communicator.

| | | |
|---|---|---|
| **Score** | This literary device is not present | **0** |
| | This literary device is present | **1** |
| | This literary device is present, with the highest rhetorical effect | **2** |

| **CF-4: Digression / Excursus**[100] | **الاستطراد** |
|---|---|

***Definition***

The communicator mentions one topic, then leaves that topic and mentions something else which may be the opposite or contrasting to the original topic, then returns to complete the original topic.

***Example***

أنا أحب برمنغهام، إنها مدينة رائعة. كنت أعيش في لندن، لكن تكلفة المعيشة كانت باهظة للغاية، وكان سكانها دائمًا بائسين. يوجد في برمنغهام الكثير من المساحات الخضراء والناس سعداء.

I love Birmingham, it's a great city. I lived in London, but the cost of living was very high, and its residents were always miserable. Birmingham has a lot of green spaces and people are happy.

[99] A. Abd Al-Ghanī, 2011, 295-300.
[100] H. Nasif et al, 2012, 109.

| ***Explanation*** | | |
|---|---|---|
| The communicator praises Birmingham by digressing to talk of the problems of London. | | |
| **Score** | This literary device is not present | **0** |
| | This literary device is present | **1** |
| | This literary device is present, with the highest rhetorical effect | **2** |

| **CF-5: Change of Topic**[101] | | حسن التخلص |
|---|---|---|
| ***Definition*** | | |
| The communicator transitions from the introduction to the main point of the text in a smooth, connected, integrated and subtle manner. | | |
| **Score** | This literary device is not present | **0** |
| | This literary device is present | **1** |
| | This literary device is present, with the highest rhetorical effect | **2** |

| **CF-6: Finesse of Requesting**[102] | | براعة الطلب |
|---|---|---|
| ***Definition*** | | |
| The communicator conveys his desire or need without explicitly making a direct request or demand. The addressee implicitly understands what the communicator is asking for. | | |
| **Score** | This literary device is not present | **0** |
| | This literary device is present | **1** |
| | This literary device is present, with the highest rhetorical effect | **2** |

[101] H. Nasif et al, 2012, 128.
[102] H. Nasif et al, 2012, 129.

| CF-7: Pleasantness of the Ending[103] | حسن الانتهاء |
|---|---|

**_Definition_**

The communicator concludes the text in a manner that is pleasing for the addressee from a lexical viewpoint.

| Score | | |
|---|---|---|
| | This literary device is not present | 0 |
| | This literary device is present | 1 |
| | This literary device is present, with the highest rhetorical effect | 2 |

| CF-8: Finesse of the Ending[104] | براعة المقطع |
|---|---|

**_Definition_**

The communicator concludes the text in a way that links to, and summarises the objective of the text, and brings a natural and satisfying conclusion to the text, from a semantical viewpoint.

| Score | | |
|---|---|---|
| | This literary device is not present | 0 |
| | This literary device is present | 1 |
| | This literary device is present, with the highest rhetorical effect | 2 |

### _3.3.7) PART G – Miscellaneous_

| CG-1: Negative elements in the text[105] | الأساليب السلبية |
|---|---|

**_Definition_**

There are a several lexical items which detract from the eloquence and rhetorical qualities of a text. These include:

[103] H. Nasif et al, 2012, 129.
[104] H. Nasif et al, 2012, 129.
[105] H. Abdul-Raof, 2006, 23 & 240-241.

- Inkhorn terms such as *kambyūtar* for “computer” which are literal transliterations into Arabic of a foreign word.
- Catachresis, where a lexical item is used incorrectly, such as when a non-native communicator does not know the correct term in that language.
- Errors in grammar or morphology.
- Phonetic incongruity where the pronunciation is difficult or distasteful.
- Unfamiliar usage of a lexical item.

***Scoring criterion***

One mark should be deduced for each occurrence of these elements.

| | | |
|---|---|---|
| **Score** | This feature is not present | **0** |
| | This feature is present | **-1** |

# Chapter 4
# Results

To illustrate the ranges of AR LD density scores produced by the tool, a range of samples of Arabic text were taken, which intentionally had a range of a priori estimated AR LD densities. These results are summarised in the table below. Appendix 5 demonstrates the end result of the scoring process with texts hand-annotated to highlight LDs, the codes for which correspond exactly with the codes presented in Chapter 3. This transparency allows different assessors to double-check, discuss, and challenge the scores that have been awarded This will make the results more accurate.

An unexpected finding was that the AR LD density summaries show that samples B & D relied on Embellishments, while sample B relied on Figures of Speech, reflective of their respective genres.

| | A | B | C | D | E |
|---|---|---|---|---|---|
| A priori LD density estimate | Zero | Low | Intermediate | High | Very high |
| LD score | 1 | 10 | 17 | 18 | 23 |
| Morphemes | 46 | 98 | 160 | 65 | 39 |
| LD density score | 0.02174 | 0.10204 | 0.10625 | 0.27692 | 0.58974 |
| | $A_1$ $B_0$ $C_0$ / 046 | $A_2$ $B_2$ $C_6$ / 098 | $A_2$ $B_{10}$ $C_5$ / 0160 | $A_3$ $B_6$ $C_9$ / 065 | $A_6$ $B_7$ $C_{10}$ / 039 |
| A: Falafel recipe<br>B: Extract from BBC News Arabic article[106]<br>C: Extract from the 1929 Egyptian short story "*ḥadīth al-qariya*" by Maḥmud Tahir Lashīn[107] | | | D: Extract from Palestinian poem "*sajjil ana ʿarabiyy*" by Mahmoud Darwish[108]<br>E: Quran, Surah Ibrahim, Verses 1 & 2. | | |

Table 1: Results from testing the tool with five sample texts

[106] https://www.bbc.com/arabic/world-62793685
[107] https://alantologia.com/blogs/57214/
[108] mssader.com/قصيدة-سجل-انا-عربي

# Chapter 5
# Discussion

Building on the experience of tools for quantitative analysis of texts in English and Arabic, this project has created a prototype for a numerical scoring system to quantitatively evaluate the density of AR LDs in Arabic texts, as a marker of the rhetoric qualities of the text itself. It is accepted that such a tool cannot evaluate the aesthetic qualities of rhetoric, just as it is impossible for any numeric measurement to describe the splendour of the Eiffel Tower in Paris, or the majesty of the Golden Gate Bridge in San Francisco. However, by measuring the number of girders, rivels, and bolts in these structures, the amount of metal and man-hours required to construct them, and the number of annual visitors or users, we can perform extremely powerful, useful and valid comparisons between them and other metal structures such as warehouses and footbridges. The same is proposed for the measurement of AR LD density in Arabic texts.

In constructing this tool, a review of AR LDs was performed and a long-list of 95 LDs which could be detected was made. Those LDs which would yield unreliable measurements were excluded. This resulted in a final list of 84 LDs which would be sought in the Arabic text under examination. By measuring the number of times each LD appeared in the text, and awarding a score of 1 or 2 each time, a sum score of the LDs is obtained. This is divided by the number of morphemes in the text, to give the Arabic Rhetoric Literary Device Density Score for that sample of text. The tool augments the score with a breakdown of the domains of AR in which LDs have been found, and a representation of the morpheme count. The tool was tested on samples of Arabic texts, with good face validity of the results.

The present study has a number of limitations. There may be AR LDs which have not been included, especially from specialist genres such pre-Islamic poetry or the Quran. This may diminish the sensitivity of the tool. However, expanding the tool to incorporate these niche LDs may reduce the usability of the tool.

The tool draws upon pre-existing definitions of which LDs are present in AR. It is possible that there are LDs in English or other languages which have not been considered by AR scholars so far for inclusion into this Arabic corpus of knowledge, and have therefore not appeared in this tool.

One weakness of the tool is that due to the complexity of the underlying AR LD science upon which the tool is built, it may be difficult for untrained assessors to accurately score the texts. Knowledge of AR amongst even native Arab speakers is limited, beyond things like metaphor, simile and metonym. This could be addressed by developing a standardised notation system to annotate texts under investigation to mark the presence of LDs as a pre-requisite to awarding a mark. This would make the marking process an educational activity in itself, and lead to more consistent and concordant scores.

The tool has yet to be validated because there is no existing standard with which to benchmark the results against: there is no pre-existing gold-standard AR measurement system for comparison. This could be addressed by assembling a panel of experts to evaluate the AR in sample texts, and benchmarking the tool's performance against their collective judgement.

The tool increases the granularity of the overall score by documenting the distribution of marks across the three constituent domains. With the use of the electronic tools which have already been created, it would be feasible to record all the scores for all the LDs for any given text. This would yield a tremendously rich dataset of AR LDs with which to investigate the variegated use of AR.

# Chapter 6
# Conclusion

This project builds upon the past two thousand years of spoken Arabic, and the past thousand years of written AR books in particular. It advances the science of AR by proposing – for the first time – a means to grade, measure and compare the use of AR LDs in Arabic texts. This is an original and novel contribution to this field.

This project has created a working prototype tool which is capable of accurately measuring the AR LD density in Arabic texts, and presenting this data for comparative analysis. On the basis of the data presented in Chapter 4, this project has rejected the null hypothesis that it is not possible to objectively measure and compare the density of AR LDs in Arabic texts.

Existing resources on AR focus on how to use LDs to create texts with AR. The focus of this project was to address the matter from the opposite end, and to see how the existence of LDs could be detected and measured in texts. Consequently, the project proposed a novel way of re-categorising the LDs in the domain of embellishments (*'ilm al-badī'*) such that it is easier for an assessor to identify LDs in texts. The basis of this is consideration of the location of the LD in the text, rather than whether they function in a semantic or lexical manner.

By demonstrating the integration of the ARLDDMS with aspects of modern technology such as websites and QR codes, this project has demonstrated the potential to make the study of AR more accessible and more vibrant, and to rejuvenate interest in this subject amongst the current generation of Arabs and non-Arabs alike.

# Appendix 1

# Scoring proforma for the Arabic Rhetoric Literary Device Density Measurement System

| | | |
|---|---|---|
| | **Arabic Rhetoric Literary Device Density Measurement System**<br><br>Visit https://BalaghaScore.com for definitions<br><br>Online version at https://BalaghaScore.com/online | SCAN ME |

| Domain A: Word Order and Sentence Structure | | | | |
|---|---|---|---|---|
| Literary Device | Score | | Literary Device | Score |
| A-1: Reporting vs informing sentences | | | A-8: Definiteness & Indefiniteness | |
| A-2: Affirmation | | | A-9: Brevity, Verbosity & Moderation | |
| A-3: The Imperative | | | A-10: Foregrounding & Backgrounding | |
| A-4: Prohibition | | | A-11: Ellipsis | |
| A-5: The Interrogative | | | A-12: Exophora | |
| A-6: Wish | | | A-13: Use of Noun in Place of Pronoun | |
| A-7: The Vocative | | | A-14: Appropriate Style & Register | |

| Domain B: Figures of Speech | | | | |
|---|---|---|---|---|
| Literary Device | Score | | Literary Device | Score |
| B1: Simile | | | B4: Metonymy / Implicit Reference | |
| B2: Metaphor | | | B5: Hinting | |
| B3: Allegory | | | B6: Pun / Paronomasia / Double-Entendre | |

## Domain C: Embellishments

### PART A – Word Choice

| Literary Device | Score | | Literary Device | Score |
|---|---|---|---|---|
| CA 1: Meaningful Proper Nouns | | | CA-8: Equivocation | |
| CA-2: Oxymoron | | | CA-9: Adornment | |
| CA-3: Amphibology | | | CA-10: Metabole | |
| CA-4: Onomatopoeia | | | CA-11: Zeugma | |
| CA-5: Litotes | | | CA-12: Al-Istikhdām | |
| CA-6: Alliteration | | | CA-13: Epizeuxis | |
| CA-7: Palindrome | | | CA-14: Epistrophe | |

### PART B – Addressing Groups

| Literary Device | Score | | Literary Device | Score |
|---|---|---|---|---|
| CB-1: Congeries | | | CB-4: Differentiation of Similar Items | |
| CB-2: Collectiveness | | | CB-5: Epanodos | |
| CB-3: Al-Taqsīm | | | | |

### PART C – Sentence Construction

| Literary Device | Score | | Literary Device | Score |
|---|---|---|---|---|
| CC-1: Antithesis / Antonymy | | | CC-5: Head-Tail | |
| CC-2: Chiasmus / Antimetabole | | | CC-6: Similarities of the Start & Finish | |
| CC-3: Al-Jinās | | | CC-7: Parallelism | |
| CC-4: Tail-Head | | | | |

## PART D – Musicality

| Literary Device | Score | | Literary Device | Score |
|---|---|---|---|---|
| CD-1: Assonance | | | CD-5: Al-Tashrīʿ | |
| CD-2: Homeoptoton | | | CD-6: Proportioning | |
| CD-3: Concordance of the Pronunciation & Meaning | | | CD-7: Excellence of Division | |
| CD-4: Concordance of Pronunciations | | | | |

## PART E – Strengthening the Argument

| Literary Device | Score | | Literary Device | Score |
|---|---|---|---|---|
| CE-1: Integration of Imagery | | | CE-12: Observation | |
| CE-2: Stacked-up Descriptions | | | CE-13: Apostrophe | |
| CE-3: Incorporation of Proverbs | | | CE-14: Personification | |
| CE-4: Abstraction | | | CE-15: Hyperbole | |
| CE-5: Quotation | | | CE-16: Beauty of Rationale / Conceit | |
| CE-6: Hinting at the Source | | | CE-17: Asteism / Affirmed Praise | |
| CE-7: Euphemism | | | CE-18: Affirmed Dispraise | |
| CE-8: Rhetorical Shift | | | CE-19: Al-Mughāyra | |
| CE-9: Epitrope | | | CE-20: Tapinosis | |
| CE-10: Evasive Response | | | CE-21: Sarcasm | |
| CE-11: Feigned Ignorance | | | CE-22: Scholastic Approach | |

| PART F – Paragraph Construction | | | | |
|---|---|---|---|---|
| Literary Device | Score | | Literary Device | Score |
| CF-1: Multi-Genre | | | CF-5: Change of Topic | |
| CF-2: Pleasantness of the Opening | | | CF-6: Finesse of Requesting | |
| CF-3: Exordium / Finesse of Initiation | | | CF-7: Pleasantness of the Ending | |
| CF-4: Digression / Excursus | | | CF-8: Finesse of the Ending | |

| PART G – Miscellaneous | | | | |
|---|---|---|---|---|
| Literary Device | Score | | Literary Device | Score |
| CG-1: Negative elements in the text | | | | |
| Total score (A) | | | Morpheme count (B) | |

| | |
|---|---|
| **Arabic Rhetoric Literary Device Density Score**<br>Total score (A) / Morpheme count (B) | |
| **Arabic Rhetoric Literary Device Density Summary** | A B C |

# Appendix 2: Screenshots from BalaghaScore.com

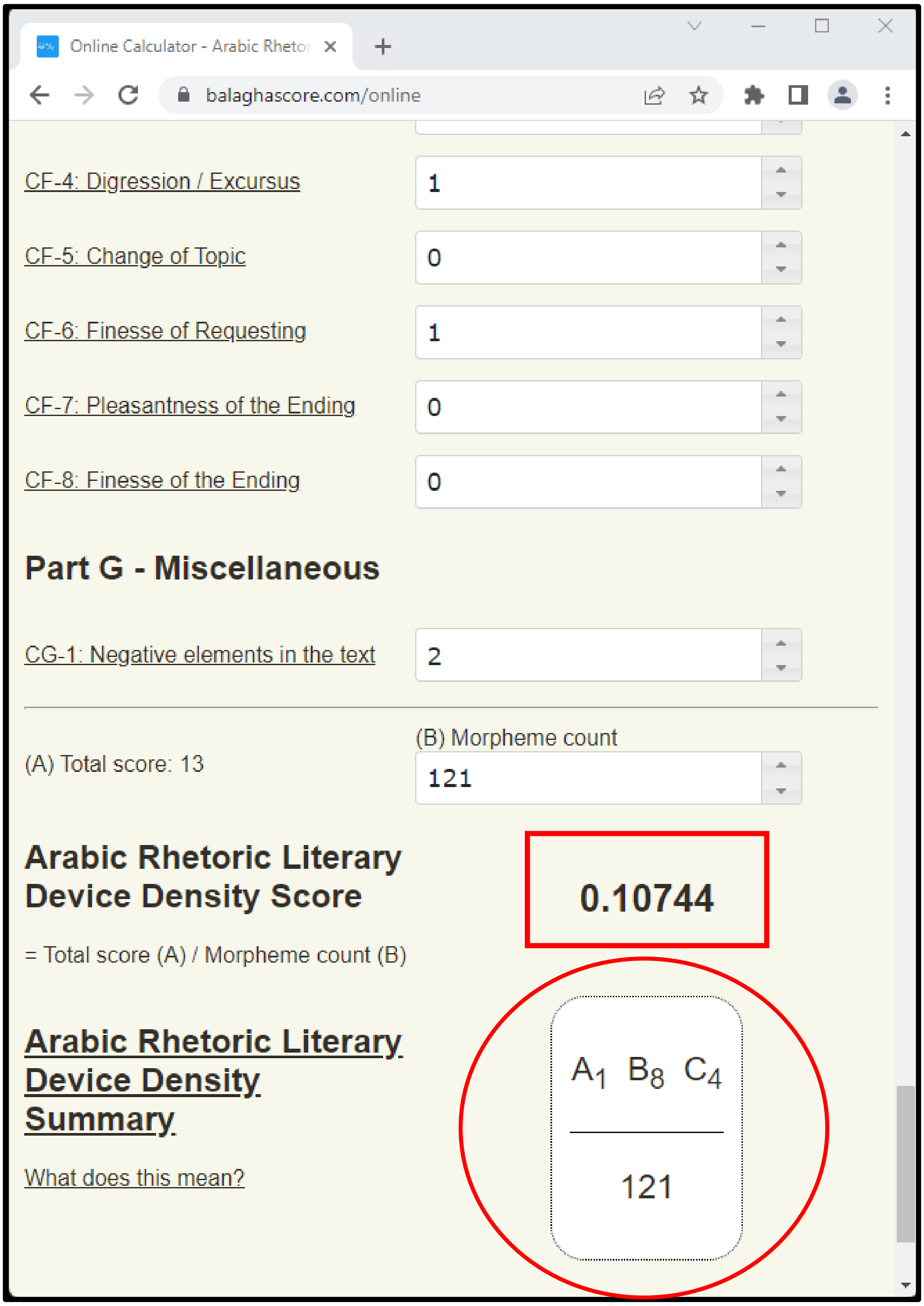

Figure 4: Example result from the online tool at https://balaghascore.com/online to calculate the AR LD density score and density summary

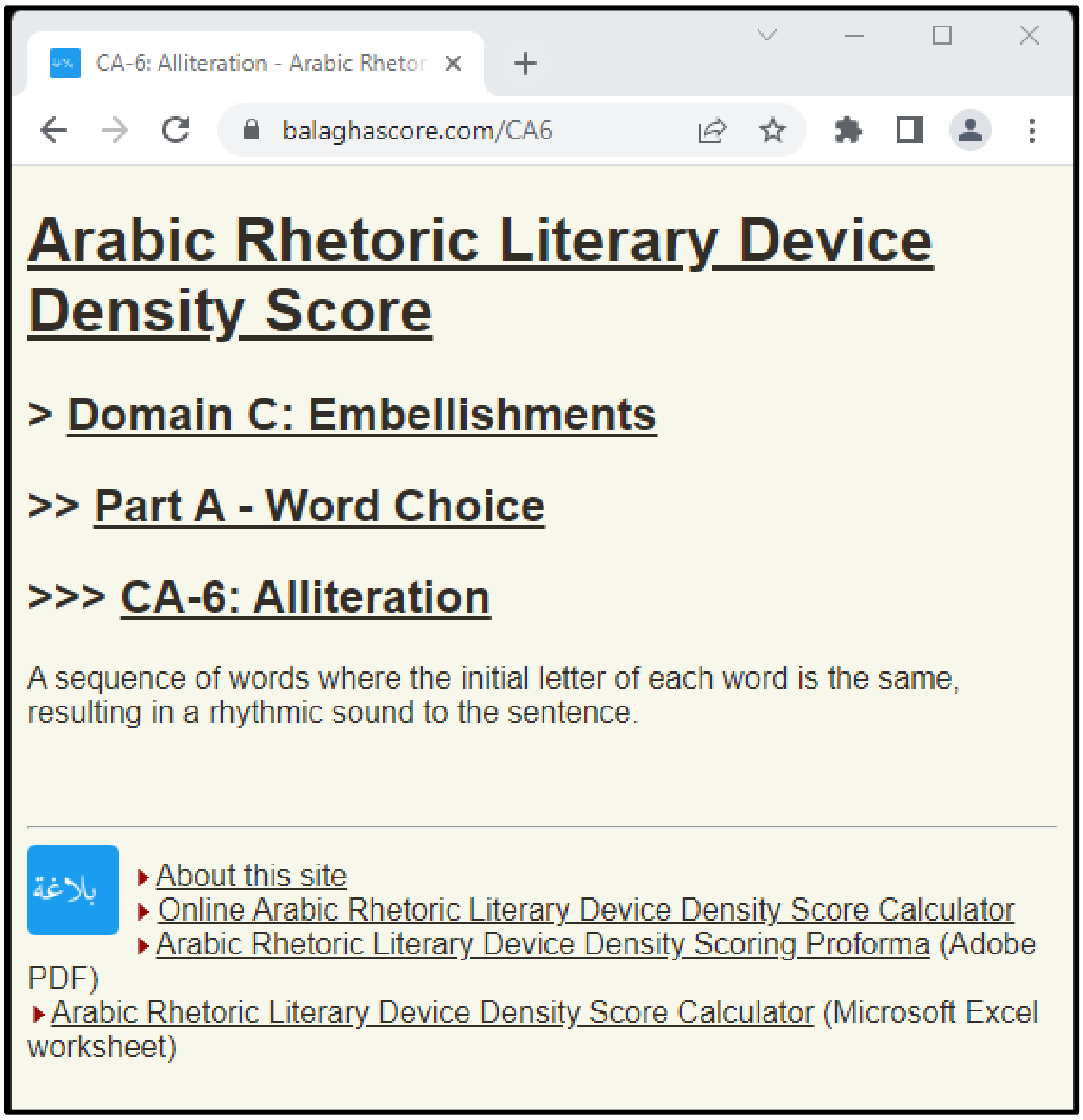

This page gives information about the definition of a LD. It can be accessed from the QR codes on the scoring proforma, the links on the Adobe PDF form, the links in the Microsoft Excel file, or from the home page of the website.

Figure 5: An example of a page on https://balaghascore.com giving information about the definition of a LD.

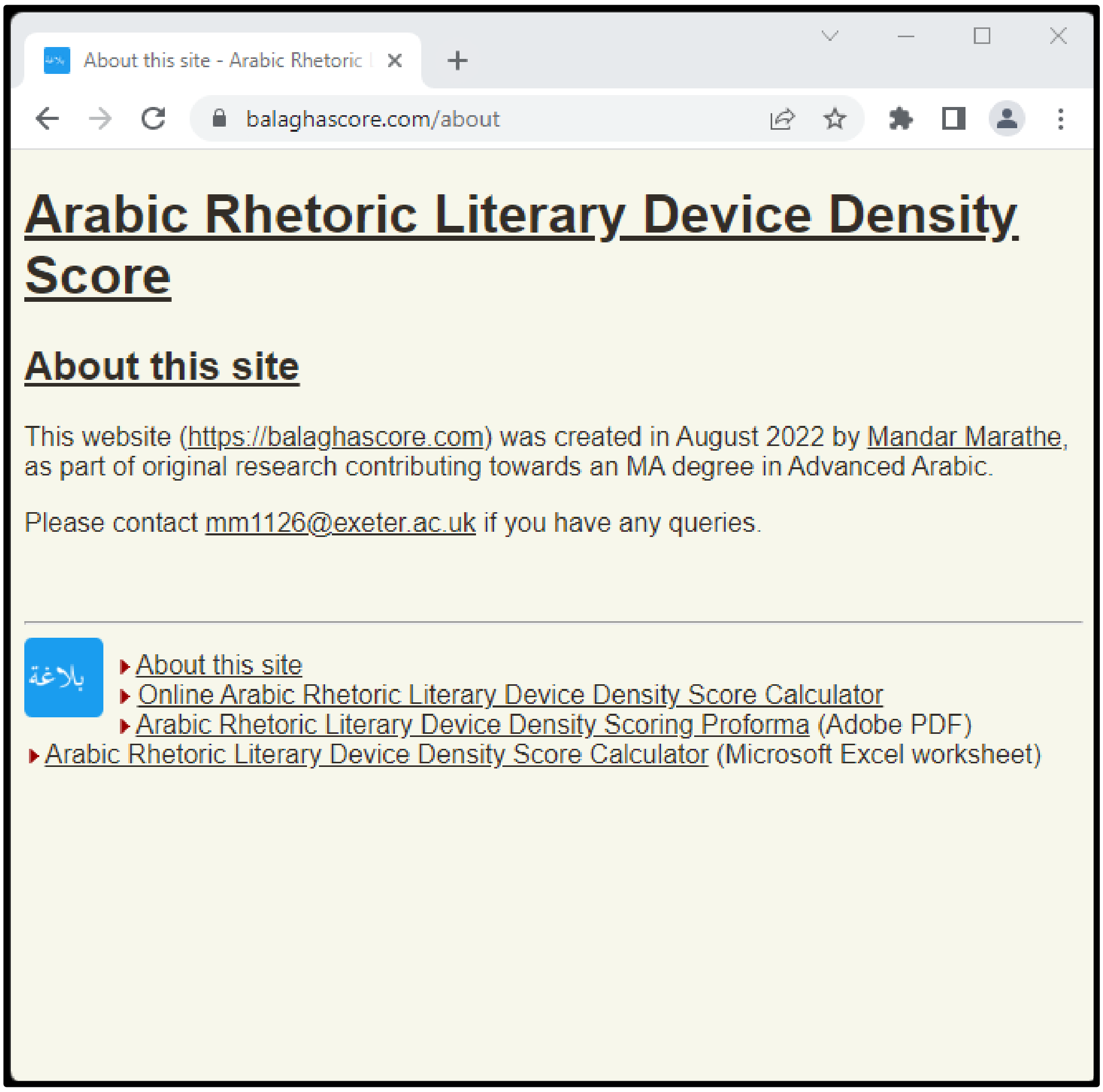

Figure 6: Screenshot from the website associated with this project (https://balaghascore.com) confirming that it is owned by the author

## Appendix 3: Screenshots from the Microsoft Excel spreadsheet which can be used to calculate scores

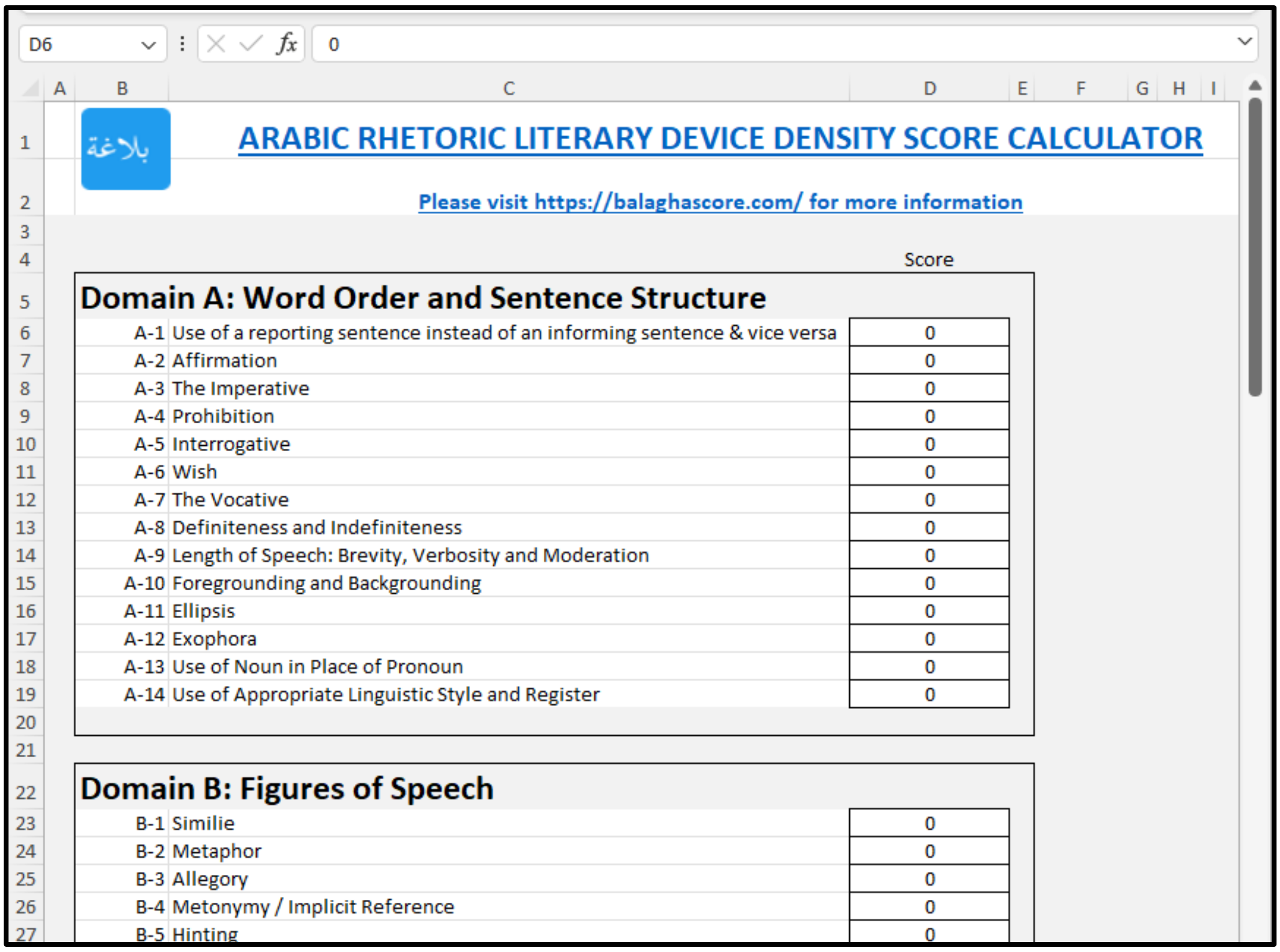

D6 | 0

**ARABIC RHETORIC LITERARY DEVICE DENSITY SCORE CALCULATOR**

بلاغة

Please visit https://balaghascore.com/ for more information

**Domain A: Word Order and Sentence Structure**

| | | Score |
|---|---|---|
| A-1 | Use of a reporting sentence instead of an informing sentence & vice versa | 0 |
| A-2 | Affirmation | 0 |
| A-3 | The Imperative | 0 |
| A-4 | Prohibition | 0 |
| A-5 | Interrogative | 0 |
| A-6 | Wish | 0 |
| A-7 | The Vocative | 0 |
| A-8 | Definiteness and Indefiniteness | 0 |
| A-9 | Length of Speech: Brevity, Verbosity and Moderation | 0 |
| A-10 | Foregrounding and Backgrounding | 0 |
| A-11 | Ellipsis | 0 |
| A-12 | Exophora | 0 |
| A-13 | Use of Noun in Place of Pronoun | 0 |
| A-14 | Use of Appropriate Linguistic Style and Register | 0 |

**Domain B: Figures of Speech**

| | | |
|---|---|---|
| B-1 | Similie | 0 |
| B-2 | Metaphor | 0 |
| B-3 | Allegory | 0 |
| B-4 | Metonymy / Implicit Reference | 0 |
| B-5 | Hinting | 0 |

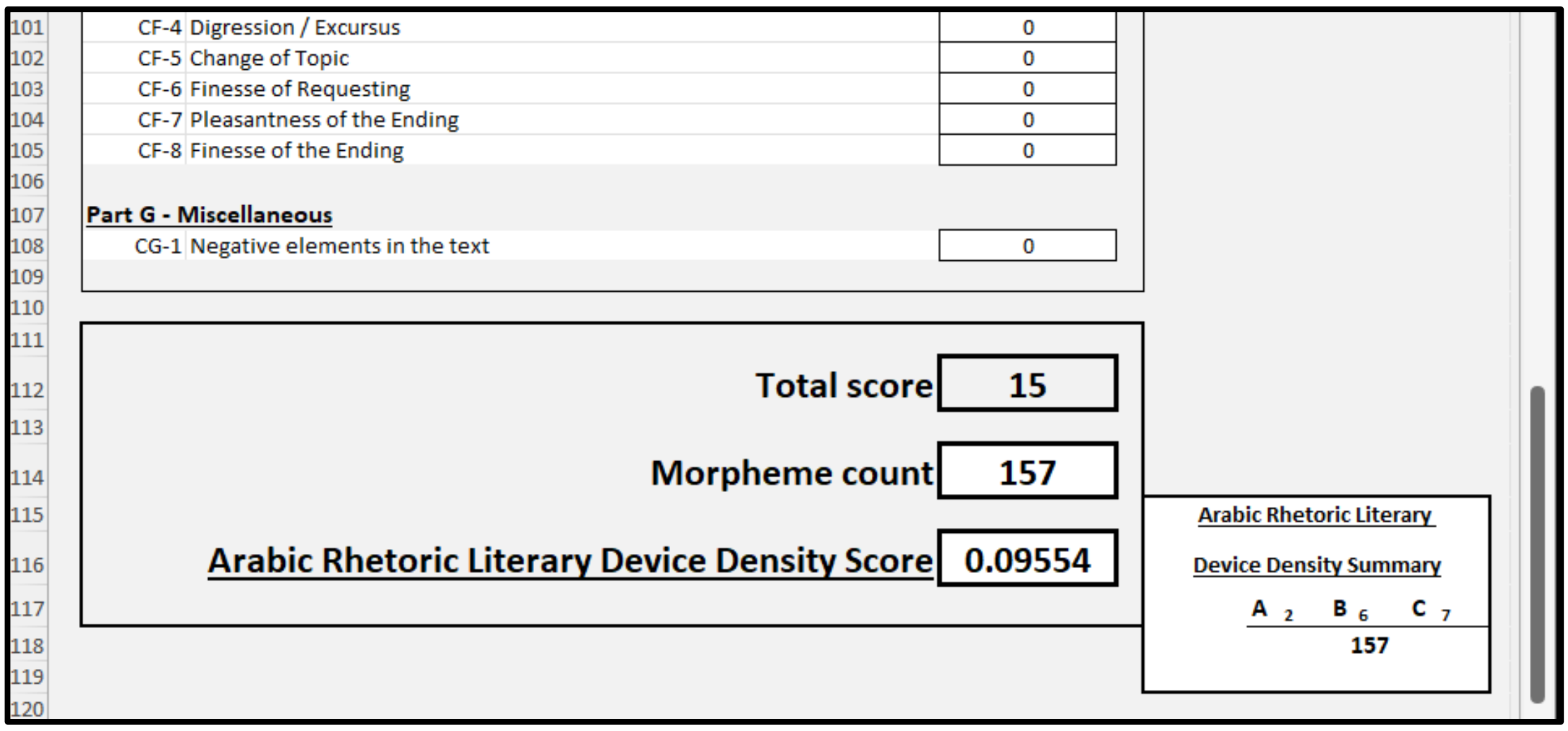

| | | |
|---|---|---|
| CF-4 | Digression / Excursus | 0 |
| CF-5 | Change of Topic | 0 |
| CF-6 | Finesse of Requesting | 0 |
| CF-7 | Pleasantness of the Ending | 0 |
| CF-8 | Finesse of the Ending | 0 |

**Part G - Miscellaneous**

| | | |
|---|---|---|
| CG-1 | Negative elements in the text | 0 |

Total score 15

Morpheme count 157

Arabic Rhetoric Literary Device Density Score 0.09554

Arabic Rhetoric Literary Device Density Summary

$\frac{A_2 \; B_6 \; C_7}{157}$

Figure 7: The scoring system as a Microsoft Excel file, which can be downloaded from https://balaghascore.com/Arabic-Rhetoric-Density-Calculator.xls

# Appendix 4: Simulation of scores

| | | Zero to Ten | | Zero to Two | |
|---|---|---|---|---|---|
| | | Text 1 | Text 2 | Text 1 | Text 2 |
| Assessor 1 | Not generous | 60 | 40 | 10 | 0 |
| Assessor 2 | | 70 | 50 | 10 | 0 |
| Assessor 3 | | 70 | 70 | 20 | 10 |
| Assessor 4 | | 90 | 60 | 10 | 0 |
| Assessor 5 | | 80 | 70 | 10 | 0 |
| Assessor 6 | | 70 | 50 | 20 | 10 |
| Assessor 7 | Generous | 100 | 40 | 10 | 0 |
| Assessor 8 | | 80 | 30 | 20 | 0 |
| Assessor 9 | | 70 | 60 | 10 | 10 |
| Assessor 10 | | 80 | 50 | 20 | 0 |
| Actual # of literary devices in text | | 10 | 5 | 10 | 5 |

Table 2: Simulation of scores with ten assessors

# Appendix 5: Samples of annotated texts

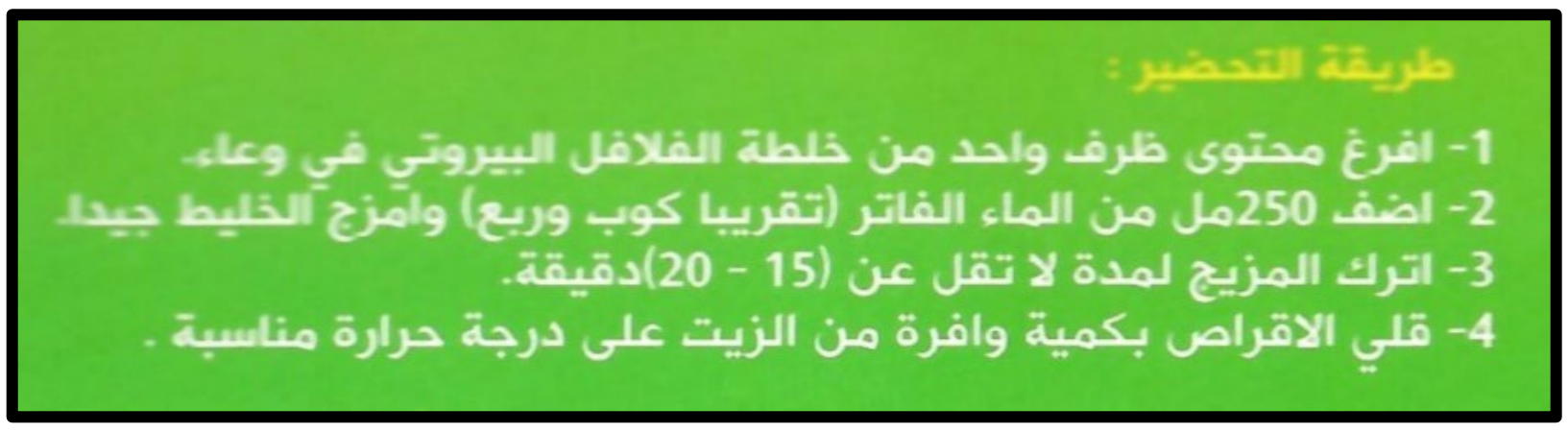

Figure 8: Text Sample A – Falafel recipie: 46 morphemes, 0 LDs

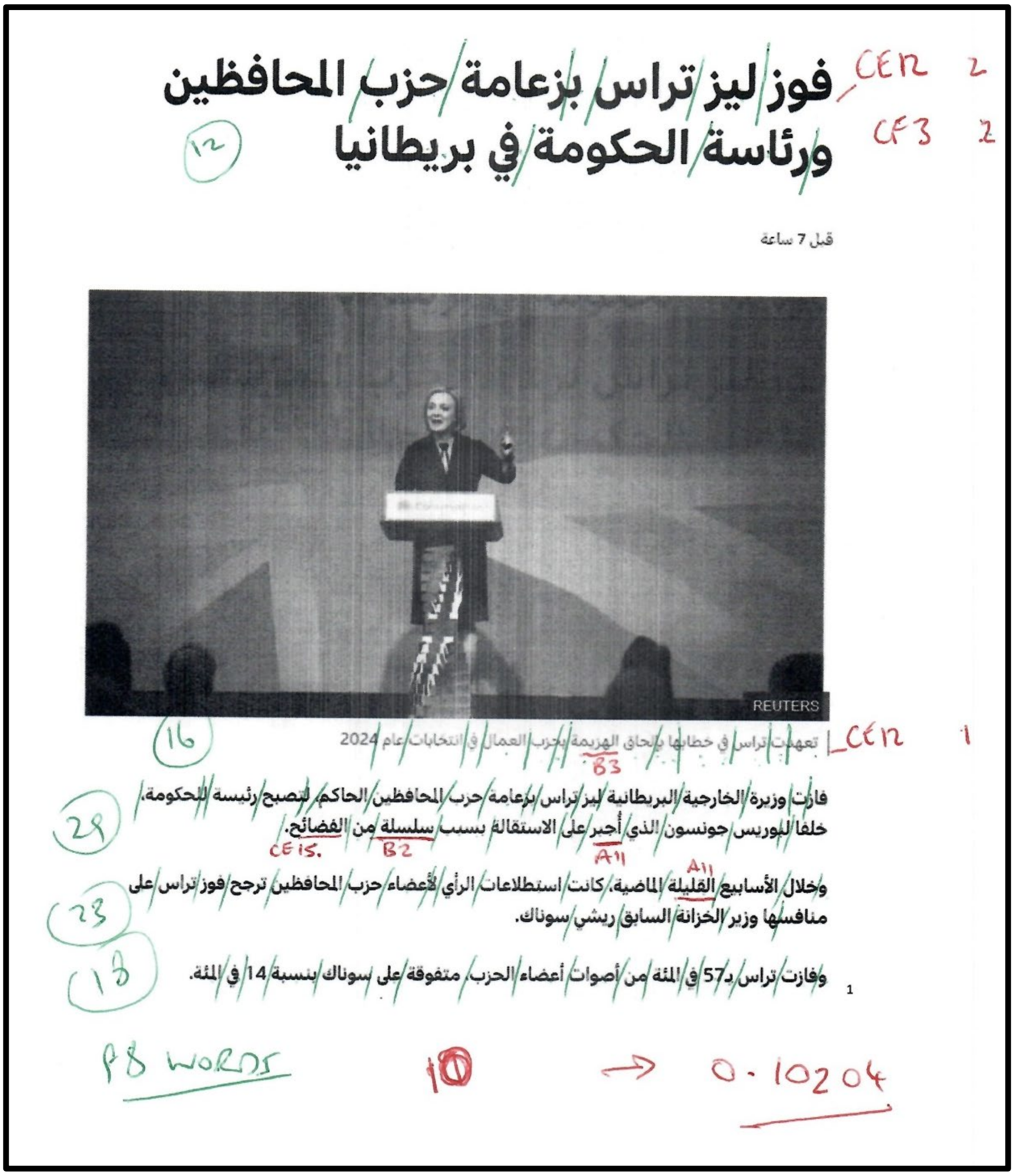

Figure 9: Text Sample B – A news article with 10 LDs and 98 morphemes

دعاني صديقي إلى أن أزور قريته معه لنقضي يوم الجمعة في أحضان
الريف وليقضي بعض مصالحه، فذهبنا، وثم لقينا نضرة وسرورا. ولكن
تلك النظرة التي تفتن ابن المدينة في لانهائية الريف، وذلك السرور الذي
يغمر كيانه ووجدانه حين يرى الطبيعة تتهلل له أيان ولى وجهه، كانا
مشوبين عندي بالرثاء للفلاحين أنصاف العرايا, وهم مكبون على
الأرض يعملون فيها الفلوس أو المناجل,مكدودين يتصببون عرقا في
أوار القيظ، وللفلاحات القابعات في ذلة لدى الأكواخ المتحذة من الطين
والبوص، وكنا إذا سرنا في الطرقات الضيقة الملتوية فأشرفنا عليهن،
تداخلن بعضهن في بعض، وتحجبن عنا بخرق بوال، حملت من تراب
الأرض بقدرما تحمل الأرض، والأولاد الصغار أنصاف عرايا كآبائهم،
قذرون كأمهاتهم، يرتعون مع الماعز والفراخ في تلك العراجين، وفوق
أكوام التراب، أو حول البركة الآسنة المحاورة.

Figure 10: Text Sample C – Short story with 17 LDs and 160 morphemes

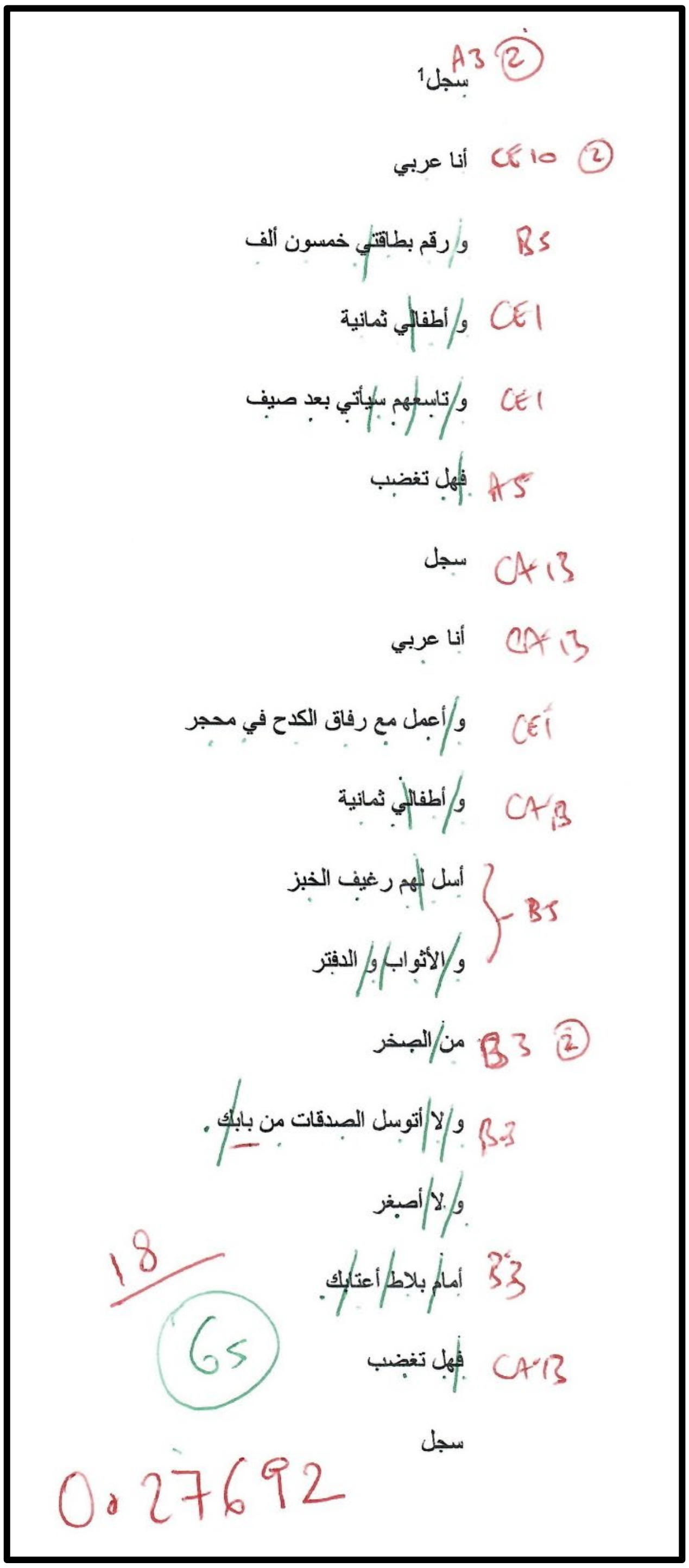

سجل[1]

أنا عربي

ورقم بطاقتي خمسون ألف

وأطفالي ثمانية

وتاسعهم سيأتي بعد صيف

فهل تغضب

سجل

أنا عربي

وأعمل مع رفاق الكدح في محجر

وأطفالي ثمانية

أسل لهم رغيف الخبز

والأثواب والدفتر

من الصخر

ولا أتوسل الصدقات من بابك

ولا أصغر

أمام بلاط أعتابك

فهل تغضب

سجل

Figure 11: Text Sample D – A famous Palestinian poem with 18 LDs and 65 morphemes

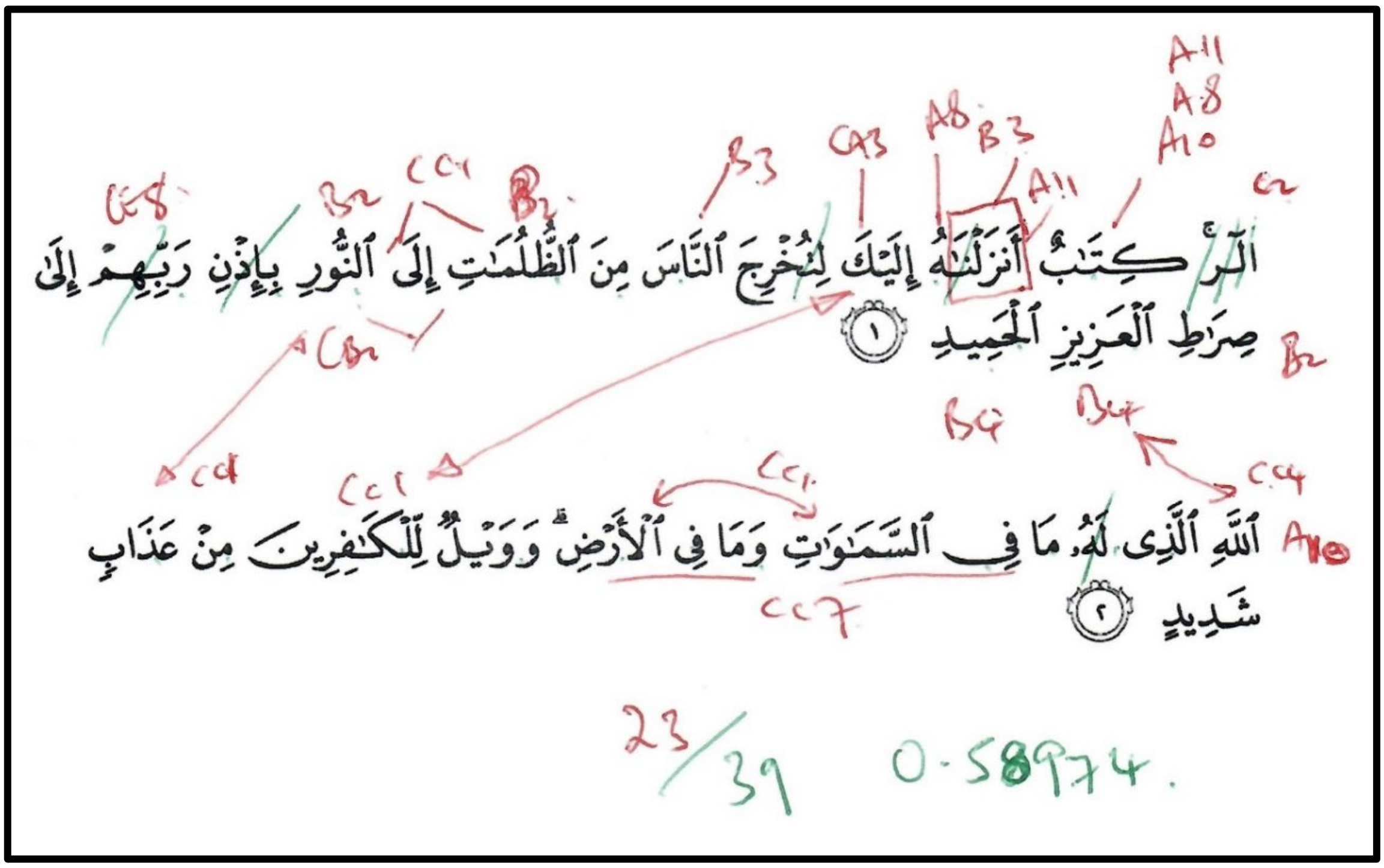

Figure 12: Text Sample E – Quran, Surah Ibrahim, Verses 1 and 2, with 23 LDs and 39 morphemes.

# Bibliography

- Abd Al-Ghani, Ayman Amin. *Al-Kafi fi Al-Balagha: Al-Bayan, Al-Badiʿ wa Al-Maʿani*. Cairo: Dar Al-Tawfiqiyya Lil-Turath, 2011.
- Abdel Haleem, Muhammad. “Rhetorical Devices and Stylistic Features of Qur’anic Grammar” in: Shah, Mustafa and Abdel Haleem, Muhammad (Eds.) *The Oxford Handbook of Qur'anic Studies* (University Press: Oxford, 2020).
- Abdul-Raof, Hussein. *Arabic Rhetoric: A Pragmatic Analysis*. Oxford: Routledge, 2006.
- Abdullah, Farhat. Zahra, Fatima. “Lexical Density and Cohesive Devices of the Divine Message in the English Translation of Surah Rehman” *International Bulletin of Literature and Linguistics* 3, 20 (2020), 38-50.
- Al-Wahy, Ahmed S. “Towards a Methodology for Measuring Lexical Density in Arabic” *International Journal of Arabic Linguistics* 3, no. 1, 1-33.
- Cavalli-Sforza, Violetta. Saddiki, Hind. Nassiri, Naoual. “Arabic Readability Research: Current State and Future Directions” *Procedia Computer Science* 142 (2018), 38-49.
- Halliday, Michael A.K and Matthiessen, Christian M.I.M. *Halliday’s Introduction to Functional Grammar* 4th Edition (New York: Routledge, 2014).
- Muhammad, Hashim. *First Steps to Understanding Balaghah*. Leicester: Al-Qalam Publications, 2016.
- Nāsif, Hafnī. Muḥammad Diab. Muṣtafā Tamum. Sultan Muḥammad. *Durūs Al-Balāgha*. Beirut: Dar ibn Ḥazm, 2012
- Ure, Jean. “Lexical density and register differentiation” in: G.E. Perren and J.L.M. Trim (Eds.), *Applications of linguistics: Selected papers of the Second International Congress of Applied Linguistics, Cambridge 1969* (Cambridge: University Press, 1971).
- Zebiri, Kate. "Towards a Rhetorical Criticism of the Qur'an." *Journal of Qur'anic Studies* 5, 2 (2003): 95-120.

## Websites

- Alantologia.com. “محمود طاهر لاشين - حديث القرية.. قصة قصيرة (1929)” Accessed September 5, 2022. https://alantologia.com/blogs/57214/
- BBC News Arabic. “فوز ليز تراس بزعامة حزب المحافظين ورئاسة الحكومة في بريطانيا.” Accessed September 5, 2022. https://www.bbc.com/arabic/world-62793685
- IJMES. "IJMES Translation and Transliteration Guide." Accessed August 29, 2022. https://www.cambridge.org/core/services/aop-file-manager/file/630d1d754cf5970011527e65/IJMES-Transliteration-Guide.pdf
- Mssader.com. “قصيدة سجل انا عربي” Accessed September 5, 2022. https://mssader.com/قصيدة-سجل-انا-عربي
- All Quranic verses were taken on https://quran.com. Accessed August 29, 2022.